\documentclass[journal]{IEEEtran}
\usepackage{enumitem}
\usepackage{times}
\usepackage{epsfig}
\usepackage{graphicx}
\usepackage{subfigure}
\usepackage{amsmath}
\usepackage{amssymb}
\usepackage{bbm}
\usepackage{booktabs}
\usepackage{color}
\usepackage{xcolor}
\usepackage{cite}
\usepackage{multirow}
\usepackage{textcomp}
\usepackage{threeparttable}
\usepackage{makecell}
\usepackage{colortbl,booktabs}
\usepackage{tablefootnote}
\usepackage{bm}
\def\cgreen{\textcolor[RGB]{0,150,0}}

\begin{document}
\title{Dual Complementary Dynamic Convolution for Image Recognition}
\author{Longbin~Yan,~\IEEEmembership{Student Member,~IEEE,}
        Yunxiao~Qin,~\IEEEmembership{Member,~IEEE,}
        Shumin~Liu,~\IEEEmembership{Member,~IEEE,}
        
        Jie~Chen,~\IEEEmembership{Senior Member,~IEEE,}
\thanks{Corresponding author: Jie Chen.}
\thanks{Longbin Yan, Shumin Liu and Jie Chen are with School of Marine Science and Technology, Northwestern Polytechnical University, Xi’an 710072, China (email: yanlongbin@mail.nwpu.edu.cn, liu$\_$shumin@ieee.org, dr.jie.chen@ieee.org). }

\thanks{Yunxiao Qin is with the Neuroscience and Intelligent Media Institute, Communication University of China, Beijing 100024, China (e-mail: qinyunxiao@cuc.edu.cn).}
}

\maketitle
\begin{abstract}
As a powerful engine, vanilla convolution has promoted huge breakthroughs in various computer tasks. However, it often suffers from sample and content agnostic problems, which limits the representation capacities of the convolutional neural networks (CNNs). In this paper, we for the first time model the scene features as a combination of the local spatial-adaptive parts owned by the individual and the global shift-invariant parts shared to all individuals, and then propose a novel two-branch dual complementary dynamic  convolution (DCDC) operator to flexibly deal with these two types of features. The DCDC operator overcomes the limitations of vanilla convolution and most existing dynamic convolutions who capture only spatial-adaptive features, and thus markedly boosts the representation capacities of CNNs. Experiments show that the DCDC operator based ResNets (DCDC-ResNets) significantly outperform vanilla ResNets and most state-of-the-art dynamic convolutional networks on image classification, as well as downstream tasks including object detection, instance and panoptic segmentation tasks, while with lower FLOPs and parameters.
\end{abstract}


\IEEEpeerreviewmaketitle

\section{Introduction}
Convolutional neural networks~\cite{lecun1998gradient} with vanilla convolution operator have been widely used in various computer vision tasks, such as image classification~\cite{krizhevsky2012imagenet,he2016deep}, object detection~\cite{ren2015faster,chen2020high,zheng2021multi} and instance segmentation~\cite{he2017mask}.
One of the most important characteristics of the vanilla convolution operator is shift-invariance, which enables the network to possess translation equivalence characteristic through local parameter sharing in the spatial domain, and thus reduce the number of parameters.
Nonetheless, as every coin has its two sides, the vanilla convolution operator suffers from three main shortcomings.
i) A single convolution kernel can only extract one single pattern in the spatial domain, which weakens the operator's representation capacity. To remedy it, hundreds of kernels are necessary for perceiving tasks in complex scenarios, which in turn increases the number of parameters ($\#$Params) and FLOPs (floating point operations).
ii) Vanilla convolution is sample-agnostic, that is, the trained kernel performs convolution operations on different inputs in inference phase with the same parameters, which is often not optimal.
iii) Vanilla convolution is not good at dealing with features of varying resolutions because the feature jitters caused by resolution changes may lead to misalignments between frozen kernel weights and features. Therefore, multi-scale training is a commonly used technique to 
improve the  CNNs' generalization ability on varying resolutions, but this measure greatly increases the training cost.

To efficiently enhance the representation capacity for diverse inputs, researchers proposed several dynamic convolution operators~\cite{sun2021gaussian,ding2021dynamic,yang2019condconv,chen2020dynamic,zhang2020dynet,li2021involution,zhou2021decoupled}. Among them, CondConv~\cite{yang2019condconv},  DynamicConv~\cite{chen2020dynamic} and Dynet~\cite{zhang2020dynet} generate multiple sets of parameters for each convolution kernel, and then weights them via an attention mechanism according to the input. This type of operator inherits the shift-invariance property of the vanilla operator while being sample-adaptive. However, they are extremely resource-intensive, especially in deep CNNs.
Involution~\cite{li2021involution} and DDF~\cite{zhou2021decoupled} tremendously alleviate these problems by adaptively dealing with diverse patterns in spatial domain. The former adopts the opposite design philosophy from vanilla convolution, where kernel parameters are adaptive in spatial domain but shared in channel domain. The latter further weights the parameters over the channels via an attention mechanism. Nevertheless, these approaches ignore the advantages of shift-invariance of vanilla convolution, which limits model's representation capacity.
This motivates us to rethink the designs of dynamic convolutional networks.
\begin{figure*}[t]
	\centering
	\includegraphics[width=16cm]{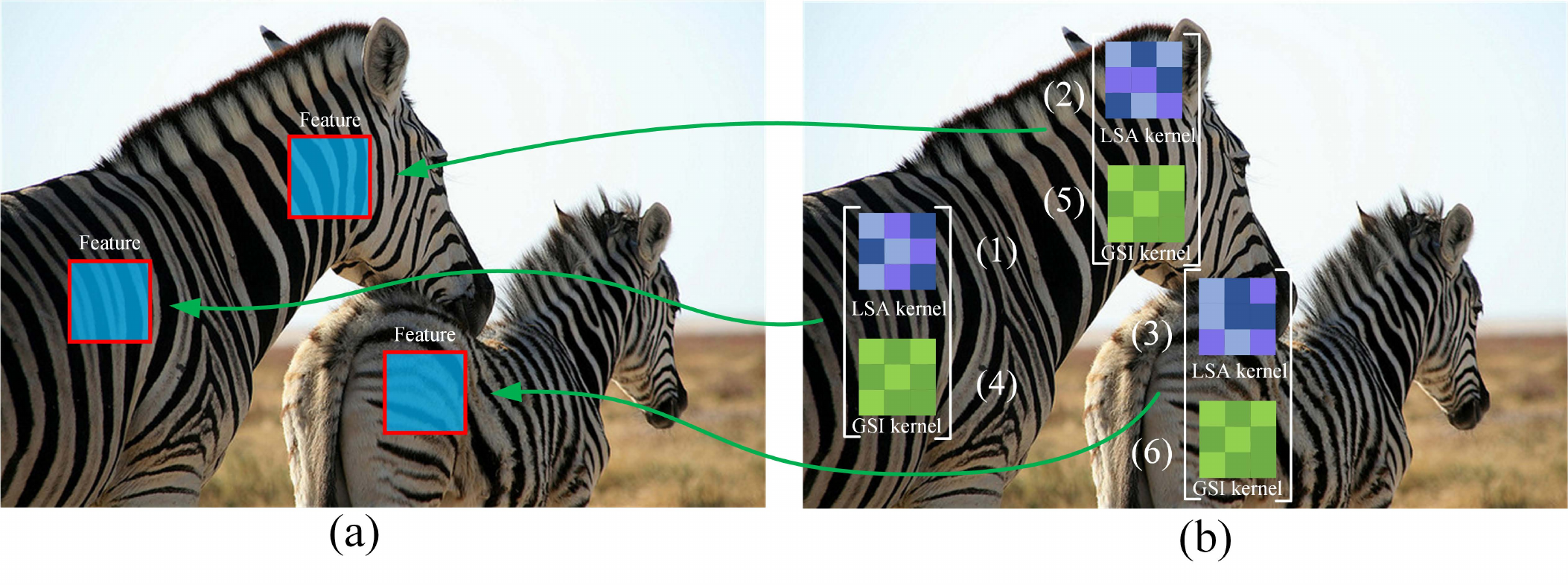}\\
	\caption{Schematic diagram of feature extraction. As shown in picture (a), the features in each window can be regarded as a combination of two parts, one is unique to individuals and another one is common to all. It motivates us to propose a two-branch dual complementary dynamic convolution operator illustrated in picture (b), which is powered with two types kernels to extract local spatial-adaptive  and  global shift-invariant features, respectively, and thus boost the representation capacity. In picture (b), (1-3) represent LSA kernels that are different in each sliding locations, and (4-6) represent GSI kernels that keep same parameters in all sliding locations.}
	\label{motivation}
\end{figure*}

Let's first look at a common features extracting scene shown in Fig.~\ref{motivation}(a). It visualizes an example where CNN is dealing with zebra stripes.
Although the features at each window locations belong to zebra stripes type, the direction, thickness, shape of each one are different.
The similar situations also occur in some dense prediction tasks, such as object detection and segmentation. We can observe that there are often both global shared features for all individuals and unique features owned by each one. Taking the scenario in Fig.~\ref{motivation}(a) to illustrate, we named them global shift-invariant (GSI) and local spatial-adaptive (LSA) features, respectively. It motivates us to design a convolutional operator illustrated in \ref{motivation}(b) that can properly extract these two types of features.

Therefore, in this paper, we propose a novel two-branch dual complementary dynamic convolutional (DCDC) operator, which aims to boost the representation capacity by simultaneously extracting both of the GSI and LSA features. In DCDC operator, the LSA branch enables the network to make specific responses to different local contents. 
The GSI branch extracts shared features of each individual or position by aggregating global information of input. 
Note that both of the LSA and GSI branches in the DCDC operator are sample-specific, which is advantageous to adapt to different input samples. Besides, benefitted from powerful modeling capabilities, DCDC operator can enable the model to reduce the number of parameters and thus alleviates overfitting.

In summary, the main contributions are: 
\begin{itemize}
\item We for the first time to model the scene features as a combination of LSA and GSI parts, and then build a DCDC operator to deal with these two types of features, which can significantly enhance the representation capacity compared with vanilla and existing state-of-the-art dynamic convolutions.
\item We propose corresponding substructures for the LSA and GSI branches to achieve a good trade-off between the parameters, FLOPs, and accuracy (see Figs.~\ref{Param_FLOPs} (a) and (b). Our proposed DCDC-ResNets can achieve a 1.9$\sim$3.8\% Top-1 accuracy gain on ImageNet-1K dataset, but only require 60$\sim$70\% of both the $\#$Params and FLOPs compared with ResNets. 
For downstream tasks such as object detection, instance and panoptic segmentation, marked improvements can also be obtained.
\end{itemize}

\section{Related Work}

\subsection{Static Convolutions}
As the core component of CNNs, convolution operator plays an important role in most computer vision tasks. 
To improve the representation capacity and reduce the computational cost of vanilla convolution, many potential variants have been proposed. 
Consideing the hundreds of filters is necessary to deal with diverse patterns, grouped convolutions~\cite{krizhevsky2012imagenet,xie2017aggregated} are used to compress features to multiple independent subspaces to reduce the number of parameters and FLOPs. 
Motivated by grouped convolutions, the more radical depth-wise separable convolutions are proposed to build light-weight deep CNNs named MobileNets~\cite{howard2017mobilenets,sandler2018mobilenetv2}, which greatly alleviates the problems of over-parameterization and heavy calculation cost caused by the excessive number of filters.
By contrast, the point-wise convolution~\cite{lin2013network} performs  mapping only across the channels with a small quantity of parameters and computational cost.
In addition, dilated convolution~\cite{yu2015multi} expands the receptive fields of vanilla convolution by inserting holes between consecutive coefficients of the filters, and thus improve the performance of computer vision tasks benefited from long-range dependence modeling, such as semantic segmentation~\cite{liang2015semantic,chen2017deeplab}.  Although static convolutions have made breakthroughs in various vision tasks, the problem of sample-agnostic weakens their abilities to adapt to different inputs.

\subsection{Dynamic Convolutions}
Due to their properties of spatial contents adaptability and input sample specificity, dynamic convolutions can make up for the above-mentioned shortcomings of static convolutions.
Common dynamic convolution can be generally classified into two categories: convolutions with dynamic structures and dynamic parameters~\cite{han2021dynamic}.
The former ones include dynamic depth~\cite{huang2018multi,bolukbasi2017adaptive}, width~\cite{eigen2013learning,bengio2015conditional}, and routing~\cite{wang2018skipnet,veit2018convolutional} types. 
However, although the dynamic structures type methods can adaptively change the inference graphs with given inputs, they often require specific training strategies, architecture designs, and careful hyper-parameters tuning~\cite{han2021dynamic}, etc.

The latter methods mainly include~\cite{ma2020weightnet,chen2020dynamic,li2020revisiting,jia2016dynamic,zhou2021decoupled,harley2017segmentation,su2019pixel,denil2013predicting,ha2016hypernetworks,li2021involution}, whose key characteristic is to adaptively predict kernel parameters according to different contents.
However, these methods usually suffer from excessive computational cost and memory consumption~\cite{jaderberg2015spatial,dai2017deformable,zhu2019deformable,zhang2020dynet,chen2020dynamic,jia2016dynamic,zamora2019adaptive},  restricted application scenes~\cite{wang2019carafe,zamora2019adaptive,gao2019lip,wang2021carafe++}, and lack the ability to model features with shift invariance characteristics~\cite{li2021involution,zhou2021decoupled}, which limit the applications of them in deep CNNs. Specifically, CondConv~\cite{yang2019condconv},  DynamicConv~\cite{chen2020dynamic}, Dynet~\cite{zhang2020dynet} and WeightNet~\cite{ma2020weightnet} aggregate multigroup predicted parameters via squeeze-and-excitation soft attention~\cite{hu2018squeeze} to obtain actual parameters for convolution operations. However, predicting several sets of expert parameters also multiplies the demand for computing resources. 
On the other side, deformable convolutional operator~\cite{dai2017deformable,zhu2019deformable} adaptively learns the grid sampling offsets and scale factors for each pixel, but this manner increases the computational cost and thus obviously damages the inference speed of the networks. 
Besides, CARAFEs~\cite{wang2019carafe,wang2021carafe++} aggregate context information to improve the up-sampling operators that originally depend on pixel interpolations. LIP~\cite{gao2019lip} combines self-attention and average pooling to obtain a pooling operator that is specific at each pixel. However, the application scenes of both CARAFEs and  LIP are relatively limited.
The more relevant methods to our work such as Involution~\cite{li2021involution} and DDF~\cite{zhou2021decoupled}, they achieve attractive performances on computer vision tasks such as image classification, object detection, and instance segmentation. However, these two methods mainly focus on constructing content-based adaptive kernels, but lack 
kernels with sufficient ability to explicitly model patterns with shifted-invariance characteristic. 

\section{Methodology}

In this section, We first briefly review the calculation rule of the vanilla convolution, and then lead to the details of the DCDC operator.

\subsection{Vanilla Convolution}

\label{Vanilla Convolution}
For a given 4-D input feature $\boldsymbol{X}\in{\mathbb{R}^{B \times C_{in} \times H \times W}}$, where $B,C_{in},H,W$ represent the sizes of mini-batch, channels, height, and width, respectively. The calculation processes of vanilla 2-D convolution with kernel weight  $\boldsymbol{\mathcal{W}}\in{\mathbb{R}^{C_{out}\times C_{in} \times k\times k}}$ and bias $\boldsymbol{\mathcal{B}}\in{\mathbb{R}^{C_{out}}}$ can be formulated as follows:
\begin{equation}
\label{convlution}
\boldsymbol{Y}_{(b,n,h,w)} = \sum\limits_{(i,j) \in {\Omega_{k \times k}}, m \in {\Omega_{in}}} {{\boldsymbol{X}_{\boldsymbol{C}_{(h,w)}^{(i,j,m)}}^{b}}{\boldsymbol{\mathcal{W}}_{(m,n,i,j)}}} + \boldsymbol{\mathcal{B}}_n,
\end{equation}
where $k$ and $C_{out}$ represent kernel size and the number of output channels, respectively. $\boldsymbol{Y}_{(b,n,h,w)}$ denotes scalar output at the $\boldsymbol{X}_{\boldsymbol{C}_{(h,w)}}^{b}$ cube for the $n$-th filter,  $\boldsymbol{X}_{\boldsymbol{C}_{(h,w)}}^{b} \in{\mathbb{R}^{C_{in} \times k \times k}}$ represents the sliding feature cube at the $(h,w)$ location of the $b$-th input sample, here 
$h \in [1,H]$,$w \in [1,W]$, and $b \in [1,B]$. 
$\Omega_{k \times k}$ represents the index set of spatial position inside kernel, it can be expressed by
\begin{equation}
\begin{aligned}
\Omega_{S} = \{ (-\left\lfloor {k/2} \right\rfloor  + i,-\left\lfloor {k/2} \right\rfloor  + j)\}, \\
i = 0,1,...,k - 1, j = 0,1,...,k - 1,
\end{aligned}
\end{equation}
where $\left\lfloor \ \right\rfloor$ denotes floor function.
$\Omega_{in}$ is the index set of input feature channels defined as
\begin{equation}
\Omega_{in} = \{ 0,1,...,C_{in}-1\}.
\end{equation}
Eq.~\ref{convlution} indicates that the vanilla convolution shares the kernel parameters in all sliding positions, which reveals its shift-invariant property. 
Besides, the vanilla convolution uses the same kernel parameters to calculate the output for different input samples, this also indicates its sample-agnostic property.
Although shift invariance characteristic enables the operator to capture same or similar features at different locations in spatial domain, it lacks content adaptability. 
Moreover, the sample-agnostic property also limits the representation ability for different input samples.

\subsection{DCDC}

As conceptually demonstrated in Fig~\ref{framework}(a), we design both of the LSA and GSI branches to extract local spatial-adaptive and global shift-invariant features, respectively, and then output the summation of them. 
It can be formulated as follow:
\begin{equation}
\begin{aligned}
\boldsymbol{Y}
&= \mathcal{F}_{lsa}(\boldsymbol X) + \mathcal{F}_{gsi}(\boldsymbol X)\\
&= \mathcal{F}_{lsa}(\boldsymbol X, \boldsymbol{\phi}(\boldsymbol X; \boldsymbol{\theta})) + \mathcal{F}_{gsi}(\boldsymbol X, \boldsymbol{\mathcal{\varphi}}(\boldsymbol X;\boldsymbol{\gamma})),\\
\end{aligned}
\end{equation}
where $\mathcal{F}_{lsa}$ and $\mathcal{F}_{gsi}$ represent convolution operations of LSA and GSI branches. $\boldsymbol{X}$ and $\boldsymbol{Y}$ denote input and output of the DCDC operator. $\boldsymbol{\mathcal{\phi}}$ and $\boldsymbol{\mathcal{\varphi}}$ represent the sub-networks used to predict the kernel parameters for LSA and GSI branches, and $\boldsymbol{\theta}$ and $\boldsymbol{\gamma}$ are the weights of $\boldsymbol{\mathcal{\phi}}$ and $\boldsymbol{\mathcal{\varphi}}$,  respectively. 
In addition, the proposed DCDC operator should be lightweight enough to be able to construct deep CNNs.

\subsubsection{LSA Dynamic Convolution Branch}

\label{SA_section}
Inspired by Involution~\cite{li2021involution}, we design the LSA dynamic convolution branch.
Instead of optimizing weights via gradient descent, the Involution operator uses an additional sub-network to directly predict kernel parameters according to local contents. The calculation processes of Involution operator with kernel size of $k_{lsa} \times k_{lsa}$ can be formulated as

\begin{equation}
\label{SA}
\boldsymbol{Y}_{(b,n,h,w)}^{lsa} = \sum\limits_{(i,j) \in {\Omega_{k_{lsa} \times k_{lsa}}}} {{\boldsymbol{X}_{\boldsymbol{C}_{(h,w,n)}^{(i,j)}}^{b}}{\boldsymbol{\mathcal{H}}_{(b,h,w)}^{(i,j)}}},
\end{equation}
\begin{equation}
\label{SA_weight}
\boldsymbol{\mathcal{H}}_{(b,h,w)} = \boldsymbol{\mathcal{\phi}} ({\boldsymbol{X}_{\boldsymbol{C}^{1 \times 1}_{(h,w)}}^{b}}),
\end{equation}
where $\boldsymbol{\mathcal{\phi}}$ represents the sub-network to adaptively predict the $k_{lsa} \times k_{lsa}$ kernel parameters relying on local contents, $\boldsymbol{\mathcal{H}}_{(b,h,w)}$ denotes the generated  parameters of $b$-th input sample at $(h,w)$ location. Superscript $lsa$ means local spatial-adaptive branch. 
The definition of the other items remain the same as Sec.~\ref{Vanilla Convolution}. 
It is observed from Eqs.~\ref{SA} and~\ref{SA_weight}  that the Involution operator possesses spatial-adaptive and sample-specific characteristics.

\begin{figure*}[t]
	\centering
	\includegraphics[width=18cm]{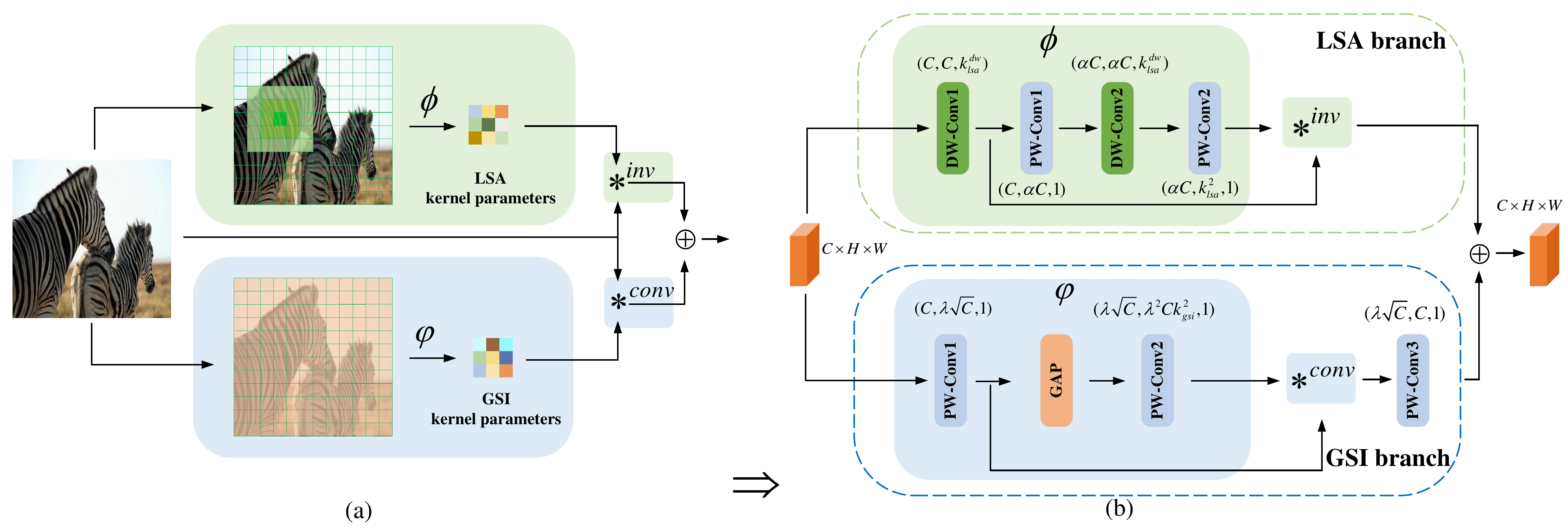}\\
	\caption{Conceptual diagram (a) and the detailed structure (b) of DCDC operator. The shaded image areas in diagram (a) represent the regions where the parameter predicting sub-networks pay attention to at a sliding location. $\boldsymbol{\phi}$ and $\boldsymbol{\varphi}$ represent the sub-networks predicting kernel parameters in LSA and GSI branches. $*^{inv}$ and $*^{conv}$ represent convolution operations of LSA and GSI branches, respectively. $(a,b,c)$-type items denote the numbers of input, output channels and kernel size of convolutional layer.}
	\label{framework}
\end{figure*}

\begin{figure*}[t]
	\centering
	\includegraphics[width=18cm]{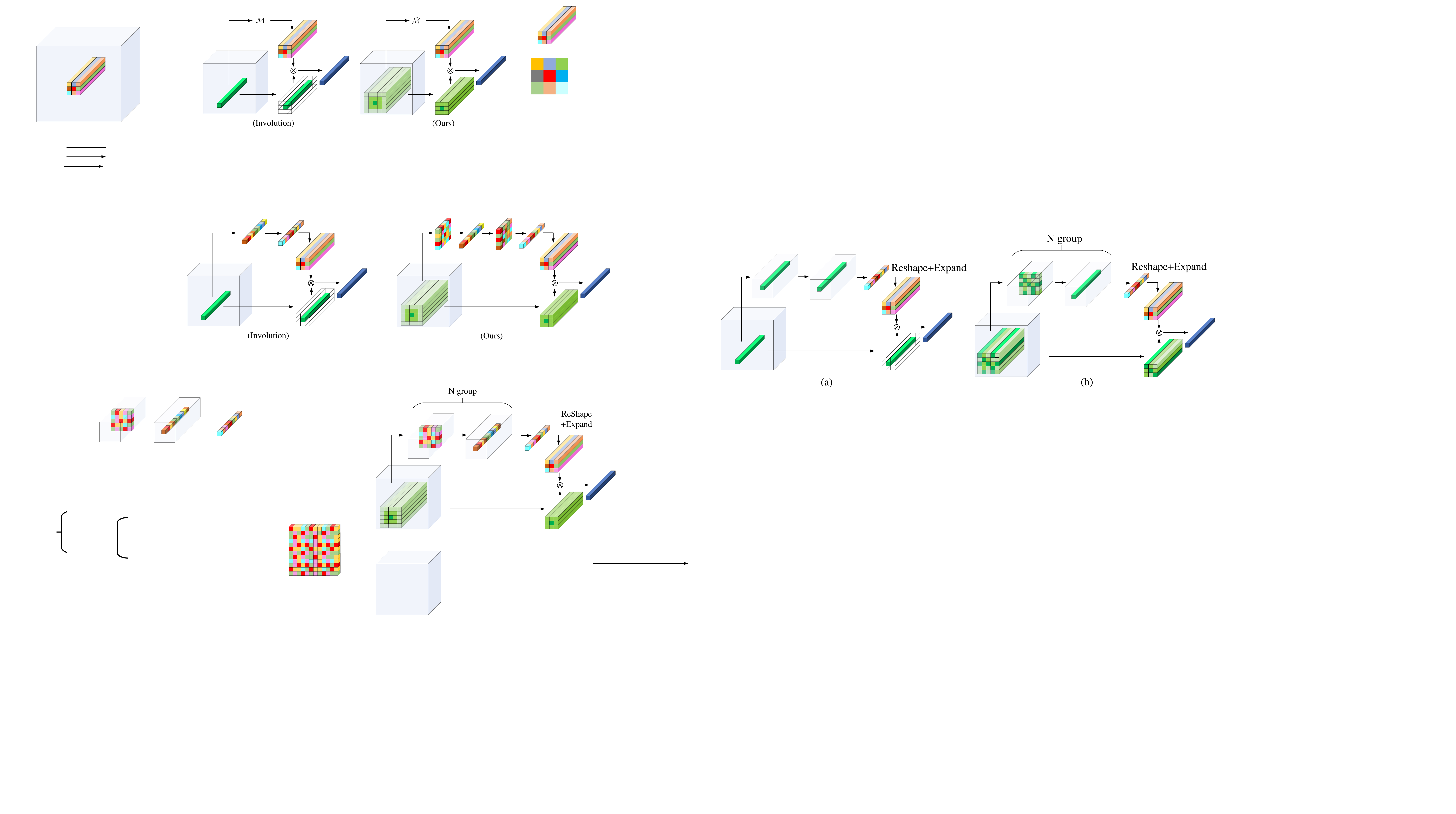}\\
	\caption{Comparisons of vanilla Involution~\cite{li2021involution} (a) and the posposed LSA operator (b). The green areas in the cubes represent the convolved pixels inside the features. \texttt{N} denotes the number of embedded depth-wise and piexel-wise convolutional layer pairs.
$\otimes$ denotes channel-wise multiplication. \texttt{Reshape} and \texttt{Expand} indicate the operations for reshaping pixel-wise weights to channel-wise ones, and then expanding weights in channel domain. Compared with Involution, the proposed LSA operator can adaptively consider pixel information in a wider range when generating $k_{lsa} \times k_{lsa}$ kernel parameters. Therefore LSA operator can achieve better performance than vanilla Involution.}
	\label{compared_with_involution}
\end{figure*}

The vanilla Involution operator shown in Fig.~\ref{compared_with_involution}(a) sets $\boldsymbol{\mathcal{\phi}}$ as two series point-wise convolutional layers and generate $\boldsymbol{\mathcal{H}}_{(b,h,w)}$ relying only on a single central pixel (denoted by superscript $1 \times 1$ in Eq.~\ref{SA_weight}) at the $(h,w)$ position. Predicting $k_{lsa} \times k_{lsa}$ kernel parameters based on only a single central pixel is obviously not optimal, it may suffer from insufficient effective receptive field. 
Note that although the center pixel on the features of deep layer already has a general receptive field much larger than a single pixel with respect to input sample, it is still limitedly benefits the prediction of 
LSA kernel parameters beyond the center pixel at current layer.
In addition, ignoring the connections between neighborhood pixels and performing multi-level mapping only on a single pixel is inefficient for modeling features.
To remedy it, as illustrated in Fig.~\ref{compared_with_involution}(b) (detailed structure shown in upper part of Fig.~\ref{framework}(b)), we skillfully embed depth-wise convolutional layers before each point-wise convolutional layer in $\boldsymbol{\mathcal{\phi}}$. It greatly alleviates these problems via expanding receptive fields on the current layer and aggregating information from neighbouring pixels. In summary, the process of predicting LSA  kernel parameters can be expressed by

\begin{equation}
\label{SA_weight_real}
\boldsymbol{\mathcal{H}}_{(b,h,w)} = \underbrace {{\sigma_{pw}}({\sigma_{dw}}}_N({\boldsymbol{X}_{\boldsymbol{C}^{s \times s}_{(h,m)}}^{b}})),
\end{equation}
where $\sigma_{dw}$ and $\sigma_{pw}$ represent depth-wise and pixel-wise convolutional layers, respectively. Here, $s \times s$ means the spatial region of pixels used to predict LSA kernel parameters, it should be at least greater than or equal to $k_{lsa} \times k_{lsa}$ in LSA branch. 
The design of alternating depth-wise and pixel-wise layers can substantially expand the
receptive fields of $\boldsymbol{\mathcal{\phi}}$ via modeling features in the spatial and channel domains without significantly increasing the computational cost. Further, this design enables the $\boldsymbol{\mathcal{\phi}}$ to generate more optimal kernel parameters, and thus improve the representation capacities of LSA branch, which is confirmed in Sec.~\ref{Ablation_experiments} (ablation experiments).

\subsubsection{GSI Dynamic Convolution Branch}
\label{SI_section}
Although vanilla convolution is shift invariant, the kernel parameters are static and not learned through focusing on global shared features, thus it may not be optimal when sliding over the whole features in spatial domain.
In addition, its sample-agnostic property also impairs the representation capacity for different input samples. To remedy this, we propose GSI dynamic convolution branch. On the one hand, it possesses shift-invariant characteristics by specifically aggregating global features to predict more appropriate parameters of GSI kernel than vanilla convolution. In other words, it has better modeling capabilities for the features shared by individuals. On the other hand, it is sample-specific that can flexibly adapt to different input samples. 
Specifically, we directly use a sub-network to predict the kernel parameters, the entire procedure of GSI dynamic convolution operator with kernel size of $k_{gsi} \times k_{gsi}$ can be defined as:
\begin{equation}
\label{SI_conv}
\boldsymbol{Y}_{(b,n,h,w)}^{gsi} = \sum\limits_{(i,j) \in {\Omega_{k_{gsi} \times k_{gsi}}}, m \in {\Omega_{in}}} {{\boldsymbol{X}_{\boldsymbol{C}_{(h,w)}^{(i,j,m)}}^{b}}{\boldsymbol{\mathcal{P}}_{(i,j,m,n)}^{b}}},
\end{equation}
\begin{equation}
\label{SI_weight}
\boldsymbol{\mathcal{P}}^{b} = \boldsymbol{\mathcal{\varphi}} ({\boldsymbol{X}^{b}}),
\end{equation}
where $\boldsymbol{\mathcal{P}}^{b} \in{\mathbb{R}^{C_{out} \times C_{in} \times k_{gsi} \times k_{gsi}}}$ represents the GSI kernel parameters predicted by sub-network $\boldsymbol{\mathcal{\varphi}}$ for the $b$-th input sample from the mini-batch. Superscript $gsi$ means global shift-invariant branch. The definition of the other items remain the same as Sec.~\ref{Vanilla Convolution}.
On the one hand, it can be observed from Eq.~\ref{SI_conv} that the GSI operator applies the same kernel to all sliding windows in spatial domain, but the different ones in channel domain. In contrast, Eq.~\ref{SA} indicates that LSA operator applies the different kernels in spatial domain, while the same one in channel domain. Therefore, both of these  information encoding approaches are dual complementary to each other, and finally enhance the performance of DCDC operator. On the other hand, it is clear from Eq.~\ref{SI_weight} that the GSI operator owns sample specificity characteristic, and thus can achieve better representation capacity for different inputs than vanilla convolution. The effectiveness of GSI branch is confirmed in Sec.~\ref{Ablation_experiments} (ablation experiments).

\textit{Implementation details.} As shown in the bottom branch of Figs.~\ref{framework}(a) and~\ref{framework}(b), the design philosophy of $\boldsymbol{\mathcal{\varphi}}$ is that it can predict more suitable GSI kernel parameters to extract global shared features only after looking features' panorama. 
Therefore, for a given feature $\boldsymbol{X}\in{\mathbb{R}^{B \times C \times H \times W}}$, we first map the channel from $C$ to $\sqrt C$ through a point-wise convolutional layer to reduce the computational cost and enhance nonlinearity. Then we aggregate the entire context information via a global average pooling layer, and use another point-wise convolutional layer to obtain the GSI kernel parameters with size of $\sqrt C \times \sqrt C \times k \times k$. 
Overall, the parameter predicting process can be formulated as follows:
\begin{equation}
\label{SI_weight_real}
\boldsymbol{\mathcal{P}}^{b} =  {\sigma_{pw}}({\text{GAP}}({\sigma_{pw}}({\boldsymbol{X}^{b}}))),
\end{equation}
where $\sigma_{pw}$ and $\text{GAP}$ represent the pixel-wise convolution and global average pooling layers, respectively. Finally, after getting the output $\boldsymbol{Y}^{gsi}$ by Eq.~\ref{SI_conv}, we use another pixel-wise convolutional layer to map channels from $\sqrt C$ to $C$ to match the original input $\boldsymbol{X}$.
\subsubsection{DCDC operator}
Here, we achieve the final DCDC operator with the formulation:
\begin{equation}
\label{DCDC}
\boldsymbol{Y}_{(b,n,h,w)} = \boldsymbol{Y}_{(b,n,h,w)}^{lsa} + \boldsymbol{Y}_{(b,n,h,w)}^{gsi}.
\end{equation}
The two terms on the right side of the equation can be obtained from Eq.~\ref{SA} and~\ref{SI_conv}, respectively. Note that the entire DCDC operator is differentiable, which facilitates us to embed it in the network and optimize end-to-end.
\section{Experiments}
Based on the proposed DCDC operator, we construct the DCDC-ResNet series backbones and train them from scratch on the ImageNet-1K dataset~\cite{deng2009imagenet}.  Then  we evaluate the transfer ability on downstream computer vision tasks including object detection, instance segmentation and panoptic segmentation on MS-COCO dataset~\cite{lin2014microsoft}.

\begin{figure*}[htbp]
\centering
\subfigure[]{
\includegraphics[width=8cm]{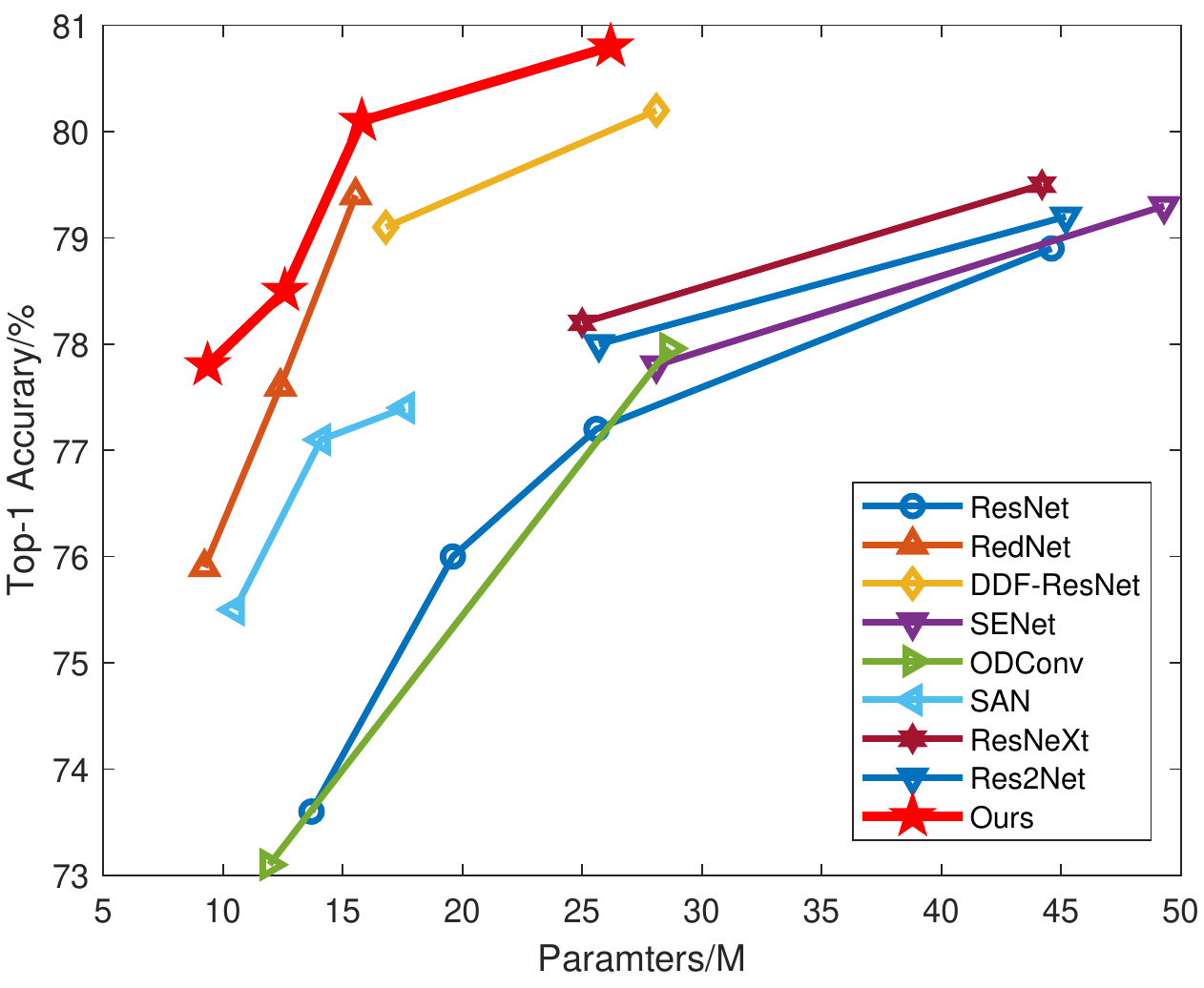}}\hspace{10mm}
\subfigure[]{
\includegraphics[width=8cm]{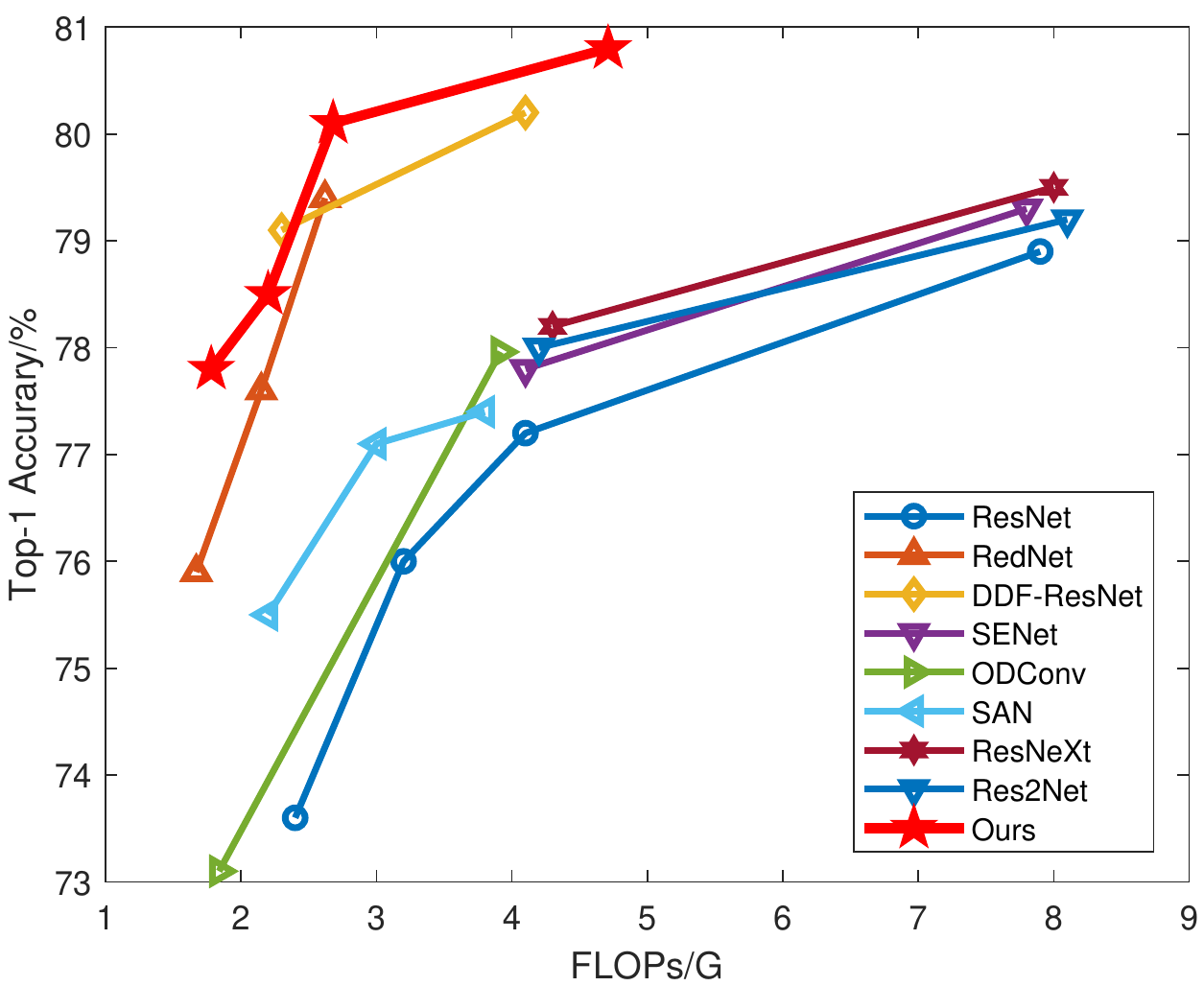}}
\caption{Comparisons of Parameters and FLOPs with existing state-of-the-art variants.}
\label{Param_FLOPs}
\end{figure*}

\subsection{Image Classification}
\subsubsection{Datasets and settings}
The ImageNet-1K dataset contains 1.28M training and 50K test images with 1K categories. Our training protocal is consistent with~\cite{li2021involution}  based on MMClassification~\cite{mmdetection} library. To construct the DCDC-ResNet series backbones, we replace 3$\times$3 convolutions in the \texttt{bottleneck} and \texttt{stem} of ResNet by DCDC operator. Considering trade-off between computational cost and accuracy, we set $k_{lsa}^{dw}$, $k_{gsi}$ and $\lambda$ hyperparameters (see Fig.~\ref{framework}(b)) of DCDC operator to 3, 1 and 1, respectively, which will be further studied in ablation experiments. $\alpha$ and $k_{lsa}$ are set to 0.25 and 7 adopted from~\cite{li2021involution}.
The mini-batch and initial learning rate with cosine annealing policy are set to 256 and 0.1. All models are trained for 130 epochs via stochastic gradient descent optimizer with the momentum of 0.9 and the weight decay of 0.0001. 
The learning rate warm-up strategy with linear growth style at first 5 epochs is also adopted to ensure the stability of convergence.
In addition, we also train the model with the same protocal with~\cite{zhou2021decoupled} based on the timm~\cite{rw2019timm} library (the training epochs is also 130) to make a fair comparison, the corresponding results are marked with * in Table~\ref{Classification_results}.

\begin{table}\small
	\renewcommand\arraystretch{1} 
	\renewcommand\tabcolsep{1pt} 
	\caption{Performance comparisons of classification experiments with other state-of-the-art methods on ImageNet-1K dataset. The highest accuracy and lowest $\#$Params and FLOPs in each set of experiments are shown in bold.}
	\label{Classification_results}    
	\begin{center}
	\begin{tabular*} {8 cm} {@{\extracolsep{\fill} }lccc} 
\toprule
Architecture & $\#$Param/M & FLOPs/G & Top-1/\%  \\
\midrule
ResNet-26~\cite{he2016deep} & 13.7 & 2.4 & 73.6  \\
LR-Net-26~\cite{hu2019local} & 14.7 & 2.6 & 75.7 \\
SA-ResNet-26~\cite{ramachandran2019stand} & 10.3 & 2.4 & 74.8 \\
SAN-10~\cite{zhao2020exploring} & 10.5 & 2.2 & 75.5 \\
RedNet-26~\cite{li2021involution} & \textbf{9.23} & \textbf{1.67} & 75.9 \\
\textbf{DCDC-ResNet-26} (\textbf{ours}) & 9.36 & 1.71 & \textbf{77.4}\\
\midrule
ResNet-38~\cite{he2016deep} & 19.6 & 3.2 & 76.0  \\
SA-ResNet-38~\cite{ramachandran2019stand} & 14.1 & 3.0 & 76.9 \\
SAN-15~\cite{zhao2020exploring} & 14.1 & 3.0 & 77.1 \\
RedNet-38~\cite{li2021involution} & \textbf{12.4} & \textbf{2.15} & 77.6 \\
\textbf{DCDC-ResNet-38} (\textbf{ours}) & 12.58 & 2.2 & \textbf{78.5}\\
\midrule
ResNet-50~\cite{he2016deep}  & 25.6 & 4.1 & 76.8/77.2*  \\
LR-Net-50~\cite{hu2019local} & 23.3 & 4.3 & 77.3 \\
CBAM-ResNet-50~\cite{woo2018cbam} &28.1&4.1& 77.3\\
AA-ResNet-50~\cite{bello2019attention} &25.8&4.2&77.7\\
SA-ResNet-50~\cite{ramachandran2019stand} & 18.0 & 3.6 & 77.6 \\
SAN-19~\cite{zhao2020exploring} & 17.6 & 3.8 & 77.4 \\
SENet-50~\cite{hu2018squeeze} &28.1&4.1& 77.8*\\
Axial ResNet-S~\cite{wang2020axial} & 12.5 & 3.3 & 78.1 \\ 
ResNeXt-50~\cite{xie2017aggregated} &25.0&4.3& 78.2*\\
Res2Net-50~\cite{gao2019res2net} &25.7&4.2& 78.0\\
DDF-ResNet-50~\cite{zhou2021decoupled} & 16.8 & 2.3 & 79.1*\\
RedNet-50~\cite{li2021involution} & 15.54 & 2.62 & 78.4/79.4* \\
\textbf{DCDC-ResNet-50} (\textbf{ours}) & 15.8 & 2.68 & \textbf{78.9}/\textbf{80.1}*\\
\midrule
ResNet-101~\cite{he2016deep} & 44.6 & 7.9 & 78.5/78.9*  \\
LR-Net-101~\cite{hu2019local} & 42.0 & 8.0 & 78.5 \\
CBAM-ResNet-101~\cite{woo2018cbam} &49.3&7.8& 78.5\\
AA-ResNet-101~\cite{bello2019attention} &45.4&8.1&78.7\\
SENet-101~\cite{hu2018squeeze} &49.3&7.8&79.3*\\
ResNeXt-101~\cite{xie2017aggregated} &44.2&8.0& 79.5*\\
Res2Net-101~\cite{gao2019res2net} &45.2&8.1&79.2 \\
RedNet-101~\cite{li2021involution} & \textbf{25.65} & 4.6 & 79.1 \\
DDF-ResNet-101~\cite{zhou2021decoupled} & 28.1 & \textbf{4.1} & 80.2* \\
\textbf{DCDC-ResNet-101} (\textbf{ours}) & 26.19 & 4.71 & \textbf{80.8}*\\
\bottomrule
\end{tabular*} 
\end{center}
\end{table}

\begin{table}\small
	\renewcommand\arraystretch{1} 
	\renewcommand\tabcolsep{1pt} 
	\caption{Performance comparisons of classification experiments with existing state-of-the-art dynamic convolution variants on ImageNet-1K dataset.}
	\label{compared_with_dynamic}    
	\begin{center}
	\begin{tabular*} {8 cm} {@{\extracolsep{\fill} }lccc} 
\toprule
Architecture & $\#$Param/M & FLOPs/G & Top-1/\%  \\
\midrule
DyNet-50~\cite{zhang2020dynet} & / &\textbf{1.1} & 76.3\\
CondConv-50~\cite{yang2019condconv} &104.8 & 4.2 & 78.6\\
MDDC-50~\cite{li2020revisiting}&30.7 & 3.9 & 77.9\\
ODConv-50~\cite{li2021omni} & 28.64 &3.9 &78.0\\
DwCondConv-50~\cite{yang2019condconv} & 14.5 & 2.3 & 78.3\\
DwWeightNet-50~\cite{ma2020weightnet} & \textbf{14.4} & 2.3 & 78.0\\
DDF-ResNet-50~\cite{zhou2021decoupled} & 16.8 & 2.3 & 79.1*\\
RedNet-50~\cite{li2021involution} & 15.54 & 2.62 & 78.4/79.4* \\
\textbf{DCDC-ResNet-50} (\textbf{ours}) & 15.8 & 2.68 & \textbf{78.9}/\textbf{80.1}*\\
\bottomrule
\end{tabular*} 
\end{center}
\end{table}

\subsubsection{Main results}
Table~\ref{Classification_results} reports the Top-1 accuracy, $\#$Params and FLOPs of DCDC-ResNet-26, -38, -50, and -101. 
It can be observed that with the same number of layers, our DCDC-ResNet backbone can consistently achieve higher accuracies while with lower FLOPs and $\#$Params compared with the seminal ResNets~\cite{he2016deep}. Especially, DCDC-ResNet-26 achieves a 3.8\% Top-1 accuracy improvement while reducing FLOPs by 29\% and $\#$Params by 32\%, compared with ResNet-26. 
DCDC-ResNet-38 even achieves comparable accuracy to ResNet-101, but only requires approximately 28\% of the FLOPs and $\#$Params.
In addition, DCDC-ResNet series has surpassed many powerful attention-based, group convolution based, etc., variants of ResNet. For example, DCDC-ResNet-50 can achieve 2.3\%, 1.9\% and 0.9\% gains in Top-1 accuracy compared with SENet-50~\cite{hu2018squeeze}, ResNeXt-50~\cite{xie2017aggregated} and Res2Net-50~\cite{gao2019res2net}, with only about 50\%$\sim$60\% $\#$Params and FLOPs. 
Note that some shallower DCDC-ResNets can even outperform most deeper networks. Such as DCDC-ResNet-38 \textit{vs.} ResNeXt-50 and Res2Net-50, DCDC-ResNet-50 \textit{vs.} ResNeXt-101 and Res2Net-101, yet DCDC-ResNets requires only about one-third computational cost. 
For presentation purposes, we also visualize the relationships between accurary, parameters and FLOPs in Figs.~\ref{Param_FLOPs} (a) and (b).

\subsubsection{Comparisons to state-of-the-art dynamic networks}
We also conduct full comparisons with stronger dynamic convolutional backbones. As shown in Table~\ref{compared_with_dynamic}, the DCDC-ResNets are still obviously superior to most state-of-the-art dynamic convolutional networks.  
Specifically, DCDC-ResNet-50 outperforms DwCondConv-50~\cite{yang2019condconv}, DwWeightNet-50~\cite{ma2020weightnet}, RedNet-50~\cite{li2021involution} and DDF-ResNet-50~\cite{zhou2021decoupled} by 0.6\% $\sim$ 2.0\% Top-1 accuracies with close computational cost. 
Note that compared with MDDC-50~\cite{li2020revisiting}, ODConv-50~\cite{li2021omni}, etc., DCDC-ResNet-50 can approximately achieve 1.0\% Top-1 accuracy, while reducing about 50\% $\#$Params and 30\% FLOPs, respectively.

\begin{table}\small
	\renewcommand\arraystretch{1} 
	\renewcommand\tabcolsep{1pt} 
	\caption{Ablation experiments of each component in DCDC operator.}
	\label{ablation_exper_component}  
	\begin{center}
		\begin{tabular*} {9 cm} {@{\extracolsep{\fill} }lccccccc} 
			\toprule
			Architecture & LSA& Conv$_\text{small}$ & Conv$_\text{large}$ &GSI& \makecell[c]{$\#$Param\\/M} & \makecell[c]{FLOPs\\/G} & \makecell[c]{Top-1\\/\%}  \\
\midrule
RedNet-26~\cite{li2021involution} &  & & & &\textbf{9.23}  & \textbf{1.67} & 75.9  \\
\midrule
DCDC-ResNet-26 & \checkmark &  & &             &9.26  & 1.69 & 76.7 \\
DCDC-ResNet-26 & \checkmark & \checkmark & &   &9.34  & 1.72 & 76.9  \\
DCDC-ResNet-26 & \checkmark &  &\checkmark &   &14.32 & 2.73 & 77.4\\
DCDC-ResNet-26 & \checkmark &  &  &\checkmark  &9.36  & 1.71 & \textbf{77.4} \\
\bottomrule
		\end{tabular*} 
	\end{center}
\end{table}

\begin{figure*}[t]
	\centering
	\includegraphics[width=18cm]{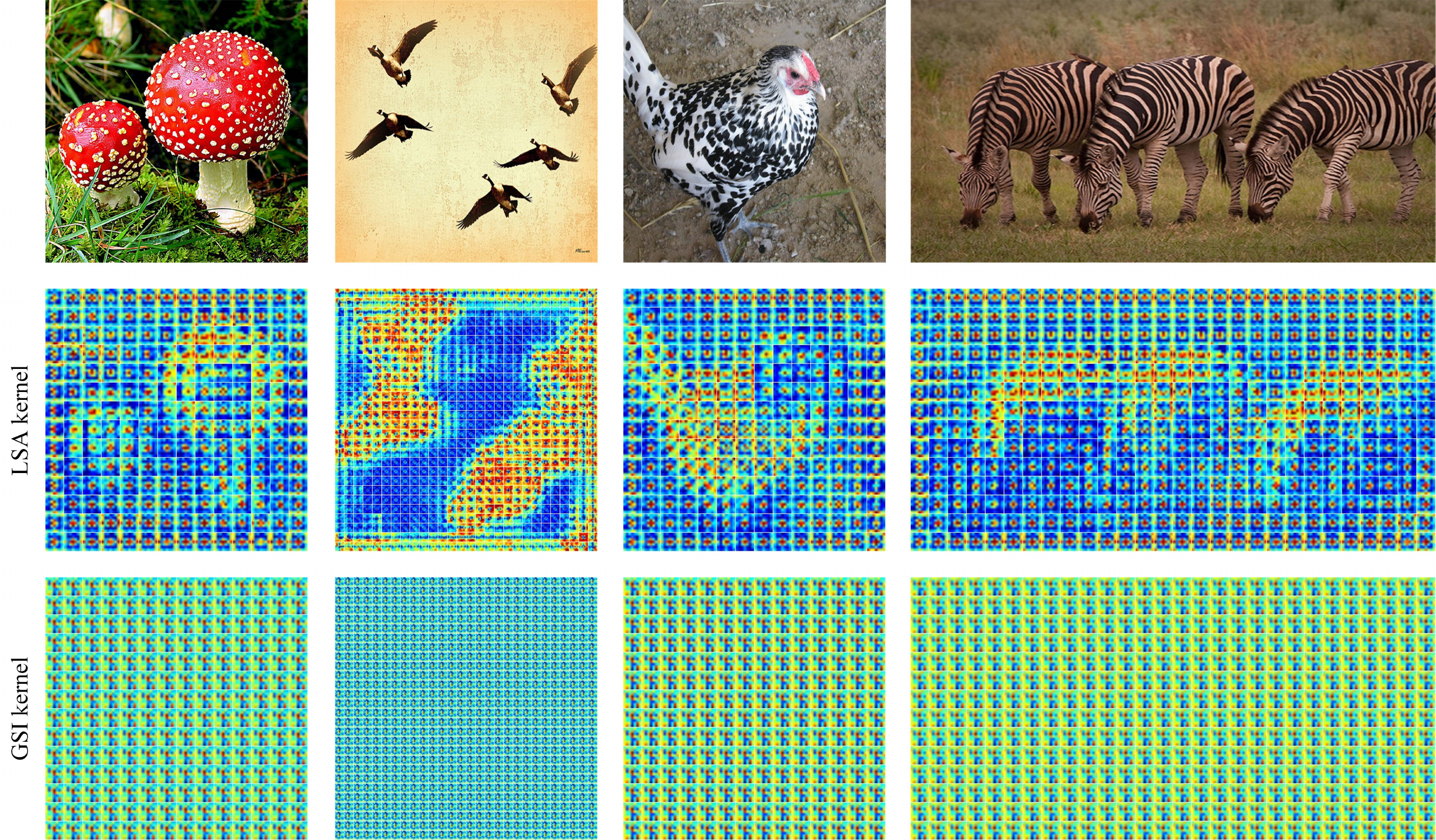}\\
	\caption{Some visualization examples of LSA and GSI kernels of the proposed DCDC-ResNet-50.}
	\label{fig_visualize_kernel}
\end{figure*}

\begin{figure*}[t]
	\centering
	\includegraphics[width=18cm]{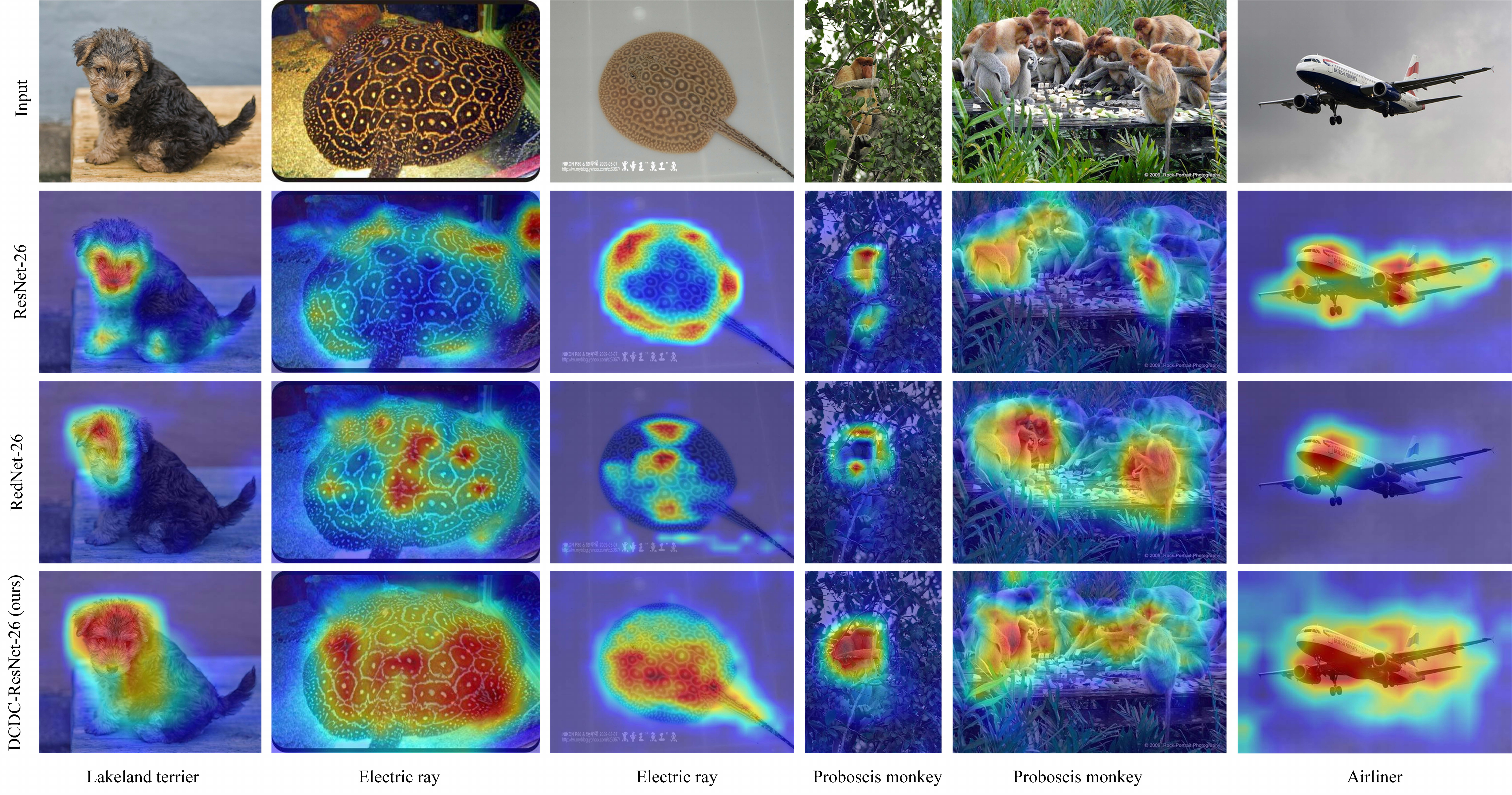}\\
	\caption{Some visualization examples of class activation maps based on Grad-Cam~\cite{selvaraju2017grad}, using ResNet-26~\cite{he2016deep}, RedNet-26~\cite{li2021involution}, and proposed DCDC-ResNet-26 backbones, respectively.}
	\label{fig_gradcam}
\end{figure*}

\subsubsection{Ablation experiments}
\label{Ablation_experiments}
In Table~\ref{ablation_exper_component}, we conduct comprehensive experiments to investigate the effectiveness of each component in DCDC operator. The baseline method is RedNet-26~\cite{li2021involution}. The LSA and GSI represents the proposed local spatial-adaptive and global shift-invariant branches. It can be seen that DCDC-ResNet only with LSA branch can achieve an obvious 0.8\% accuracy improvement with negligible increases of $\#$Params and FLOPs compared to vanilla RedNet~\cite{li2021involution}. This proves that predicting kernel parameters based on large-scale information in LSA branch is significantly better than that based on single-pixel regions as discussed in Sec.~\ref{SA_section} and Fig.~\ref{compared_with_involution}(b).

we design two contrast experiments to meticulously verify the effectiveness of the GSI branch. One experiment is carried out by replacing the dynamic convolution (i.e., \texttt{GAP} $\rightarrow$ \texttt{PW-conv2} $\rightarrow$ $*^{conv}$ in Fig.~\ref{framework}(b)) in GSI branch with a 3$\times$3 vanilla convolution layer, and the other one is replacing the entire GSI branch with a 3$\times$3 vanilla convolution layer (this one has more feature channels than prior), which are denoted as Conv$_\text{{small}}$ and Conv$_\text{{large}}$, respectively. The former can only get a slight 0.2\% accuracy increase. Although the latter can obtain a 0.7\% accuracy gain, the corresponding $\#$Params and FLOPs are risen significantly. In contrast, the experimental result with GSI branch can achieve a 0.7\% accuracy improvement while with only insignificant increases in $\#$Params and FLOPs. These experiments solidly demonstrate the effectiveness of the GSI branch.

\begin{table}\small
	\renewcommand\arraystretch{1} 
	\renewcommand\tabcolsep{1pt} 
	\caption{Ablation experiments of variates $k_{lsa}$, $k_{gsi}$ and $\lambda$ in LSA and GSI branch.}
	\label{ablation_exper_variate}   
	\begin{center}
		\begin{tabular*} {8 cm} {@{\extracolsep{\fill} }lcccccc} 
\toprule
Architecture &$k_{lsa}^{dw}$ & $k_{gsi}$ & $\lambda$ & \makecell[c]{$\#$Param\\/M} & \makecell[c]{FLOPs\\/G} &\makecell[c]{Top-1\\/\%}  \\
\midrule
DCDC-ResNet-26  & 3$\times$3&1$\times$1&1.0  &\textbf{9.36}& \textbf{1.71} & 77.38 \\
\midrule
DCDC-ResNet-26 & 5$\times$5&1$\times$1&1.0   &9.39 & 1.73 & 77.55\\
DCDC-ResNet-26 & 7$\times$7&1$\times$1&1.0 &  9.45 & 1.76 & 77.63\\
\midrule
DCDC-ResNet-26 &3$\times$3 & 3$\times$3 &1.0 & 9.61 & 1.71 & 77.39  \\
DCDC-ResNet-26 &3$\times$3 & 5$\times$5 &1.0 & 10.12& 1.71 & 77.40 \\
\midrule
DCDC-ResNet-26 &3$\times$3 &1$\times$1 & 1.5 &9.46 & 1.72 & 77.53  \\
DCDC-ResNet-26 &3$\times$3 &1$\times$1 & 2.0 &9.64 & 1.73 & 77.67  \\
\midrule
DCDC-ResNet-26 &7$\times$7 &1$\times$1 & 2.0 &9.74  &1.78 & 77.80  \\
DCDC-ResNet-26 &3$\times$3 &7$\times$7 & 2.0 &21.88 & 1.74 & \textbf{77.90} \\
\bottomrule
		\end{tabular*} 
	\end{center}
\end{table}

In Table~\ref{ablation_exper_variate}, we study the effects of (a) the kernel size ($k_{lsa}^{dw}$) of the embedded depth-wise convolutions in LSA branch, (b) the kernel size ($k_{gsi}$) of the designed dynamic convolution in GSI branch, and (c) the scaling factor $\lambda$ controlling the number of feature channels in GSI branch on the experimental performance. Here, the baseline method is DCDC-ResNet-26, the corresponding parameters $k_{lsa}^{dw}$, $k_{gsi}$, and $\lambda$ are set to $3\times3$, $1\times1$, and 1.0, respectively. 

(a) Increase $k_{lsa}^{dw}$ while keeping $k_{gsi}$ and $\lambda$ unchanged.  Results show that the performance of network can be significantly improved, while the growth of $\#$Params and FLOPs is almost negligible. 

(b) Increase $k_{gsi}$ while keeping $k_{lsa}^{dw}$ and $\lambda$ unchanged.  Results show that the performance gains are marginal.

(c) Increase $\lambda$ while keeping $k_{lsa}^{dw}$ and $k_{gsi}$ unchanged. Results show that the performance can be greatly improved with marginal computational cost increase.

Finally, after the above analyses, we comprehensively adjust $k_{lsa}^{dw}$, $k_{gsi}$, and $\lambda$ to obtain a DCDC-Resnet-26 with 77.8\% Top-1 accuracy, extremely low $\#$Params (9.74 M) and FLOPs (1.78 G), which exceeds all other 38-layer and some 50-layer networks in Table~\ref{Classification_results}.  This fully demonstrates the efficiency of DCDC-ResNets.

\subsubsection{Visualizations of LSA and GSI Kernels} In this subsection, to study the modeling ability of the DCDC operator for local spatial-adaptive and global shift-invariant features, we visualize the LSA and GSI kernels, respectively. In Fig.~\ref{fig_visualize_kernel}, we randomly choose some images from the ImageNet dataset evaluation split to visualize the LSA and GSI kernels of 2nd-stage or 3rd-stage of DCDC-ResNet-50. It can be observed that the kernels of the LSA branch own a strong contents-adaptive characteristic, and can sketchily perceive the contour of the objects. Moreover, we can also observe that when encountering features with similar patterns, the LSA branch can generate similar kernels. Because of the consistent response to similar patterns, the network can focus on a larger range rather than some specific local features when predicting the results. This point is also confirmed again in the next subsection, i.e., Sec.~\ref{CAM_section}. 
In contrast, the kernel of GSI branch is global shift-invariant, which can aggregate the global shared features. Note that different from vanilla convolution, the GSI dynamic operator is sample-specific and has more strong ability to deal with different input samples.

\subsubsection{Class Activation Maps}
\label{CAM_section}
To further understand the modeling ability of DCDC operator, we use Grad-Cam~\cite{selvaraju2017grad} to visualize the class activation map (CAM), which is a commonly used technique to reveal the regions of interest for CNNs to make predictions. As shown in Fig.~\ref{fig_gradcam}, we randomly choose some images to generate the CAMs, which have a redder color in areas with stronger attention. It can be observed that compared to ResNet, the proposed DCDC-ResNet has more consistent and stronger modeling ability for local adaptive features that contain discriminative information of the object, such as texture features on body of `electric ray'. Benefited from GSI branch and LSA branch with larger receptive field, DCDC-ResNet tends to focus on the whole of the object rather than some specific local features compared with RedNet, such as objects of `lakeland terrier', `electric ray' and `airliner', etc., which enables the network to make more accurate predictions. In addition, DCDC-ResNet can more precisely capture features to recognize hidden objects in complex environments compared with ResNet and RedNet, such as `monkey' in trees.

\renewcommand{\arraystretch}{1.0}
\setlength\tabcolsep{1pt}
\begin{table*}\small
\caption{Performance comparisons of object detection experiments on MS COCO dataset. Neck means the feature pyramid network in detectors, here, it constructed based on standard convolution, Involution~\cite{li2021involution}, and proposed DCDC operators, respectively. The highest accuracy, lowest $\#$Params and FLOPs in each set of experiments are shown in bold. \bm{$\Delta$} represents the absolute improvements in AP and the percentage reductions in $\#$Params and FLOPs.}
\label{Det_results}
\begin{center}
\begin{tabular*}{18 cm} {@{\extracolsep{\fill} }ccccccccc}
\toprule
Detector & Backbone &Neck & $\#$Param/M  & FLOPs/G &AP$_\text{bbox}$ & AP$_\text{S\_bbox}$ & AP$_\text{M\_bbox}$ & AP$_\text{L\_bbox}$\\
\midrule
\multirow{9}{1cm}{Retina-Net~\cite{lin2017focal} }  
&\textbf{DCDC-ResNet26} & \textbf{DCDC} &20.2 &179.4&37.3  & 20.6 & 41.4 & 49.2\\
\cmidrule(lr){2-9}
&ResNet50~\cite{he2016deep} (\textit{baseline}) & Convolution & 37.7 &239.3& 36.6 & 20.9& 40.6 &47.5\\
 &RedNet50~\cite{li2021involution} & Convolution & 27.8 &208.6& 38.3&  21.1& 41.8 &50.9\\
 &\textbf{DCDC-ResNet50} & Convolution &28.1 &210.0& \textbf{39.6}  &22.7 &\textbf{43.5} &53.2 \\
 &RedNet50~\cite{li2021involution} & Involution~\cite{li2021involution} & \textbf{26.3} &\textbf{197.7}& 38.2& 21.8& 41.6 &50.7\\
&\textbf{DCDC-ResNet50} & \textbf{DCDC} & 26.5 &199.3& 39.5 &\textbf{23.3}&43.2&\textbf{52.4}\\
  &\bm{$\Delta$} & & \cgreen{$\downarrow$ 30\%} & \cgreen{$\downarrow$ 17\%} & \cgreen{+2.9} &\cgreen{+2.4} & \cgreen{+2.6} & \cgreen{+4.9} \\
\cmidrule(lr){2-9}
&ResNet101~\cite{he2016deep}  & Convolution & 56.7 & 315.4 & 38.5 & 21.7& 42.8& 50.4\\

\midrule
\multirow{9}{1cm}{Faster-RCNN~\cite{ren2015faster}}  
&\textbf{DCDC-ResNet26} & \textbf{DCDC} & 23.5 &114.5 &38.7 & 22.4 & 42.2 & 50.8\\
\cmidrule(lr){2-9}
&ResNet50~\cite{he2016deep} (\textit{baseline}) & Convolution & 41.5 &207.1& 37.7& 21.7& 41.6 &48.4\\
&RedNet50~\cite{li2021involution} & Convolution & 31.6 &176.4& 39.5& 23.3& 42.9 &52.2\\
 &\textbf{DCDC-ResNet50} & Convolution &31.9 &177.7&40.4 & 24.0&43.8&52.6\\
 &RedNet50~\cite{li2021involution} & Involution~\cite{li2021involution} & \textbf{29.5} &\textbf{132.0}& 40.2 & 24.2& 43.3 &52.7\\
&\textbf{DCDC-ResNet50} & \textbf{DCDC} &29.8 & 134.4&\textbf{40.9} &\textbf{24.6}&\textbf{44.1}&\textbf{53.5}\\
 &\bm{$\Delta$} & & \cgreen{$\downarrow$ 28\%} & \cgreen{$\downarrow$ 35\%} & \cgreen{+3.2} &\cgreen{+2.9} & \cgreen{+2.5} & \cgreen{+5.1} \\
 \cmidrule(lr){2-9}
&ResNet101~\cite{he2016deep}  & Convolution & 60.5 & 283.1 & 39.4 & 22.4& 43.7& 51.1\\
\bottomrule
\end{tabular*}
\end{center}
\end{table*}

\renewcommand{\arraystretch}{1.0}
\setlength\tabcolsep{1pt}
\begin{table*}\small
\caption{Performance comparisons of instance segmentation experiments on MS COCO dataset.}
\label{instance_seg_results}
\begin{center}
\begin{tabular*}{18 cm} {@{\extracolsep{\fill} }ccccccccc}
\toprule
Detector & Backbone &Neck & $\#$Param/M  & FLOPs/G &AP$_\text{bbox}$/AP$_\text{seg}$ & AP$_\text{S\_bbox/seg}$ & AP$_\text{M\_bbox/seg}$ & AP$_\text{L\_bbox/seg}$\\
\midrule
\multirow{9}{1cm}{Mask-RCNN~\cite{he2017mask}} 
&\textbf{DCDC-ResNet26} & \textbf{DCDC} &26.1 &167.6&39.5/35.4 &22.8/16.7 &43.2/38.1 &51.7/51.6\\ 
\cmidrule(lr){2-9}
&ResNet50~\cite{he2016deep} (\textit{baseline}) & Convolution & 44.2  &260.1&38.4/35.1 &21.9/18.5 &42.3/38.6 &49.7/46.9\\ 
&RedNet-50~\cite{li2021involution} & Convolution & 34.2  &229.5&40.2/36.1 &24.2/19.9 &43.4/39.3 &52.5/48.9 \\
&\textbf{DCDC-ResNet50} & Convolution & 34.5& 230.8&40.9/36.5 &23.9/17.4 &44.4/39.5 &52.8/52.2 \\
&RedNet-50~\cite{li2021involution} & Involution~\cite{li2021involution} & 32.2  &185.1&40.8/36.4 &24.2/\textbf{19.9} & 44.0/39.4&53.0/49.1 \\

&\textbf{DCDC-ResNet50} & \textbf{DCDC} &32.5 &187.4&\textbf{41.7}/\textbf{37.1}&\textbf{25.8}/19.1 &\textbf{45.1}/\textbf{39.9} &\textbf{53.7}/\textbf{53.0}\\

&\bm{$\Delta$} & &\cgreen{$\downarrow$ 26\%} & \cgreen{$\downarrow$ 28\%} &\cgreen{+3.3}/\cgreen{+2.0} &\cgreen{+3.9}/\cgreen{+0.6}&\cgreen{+2.8}/\cgreen{+1.3}&\cgreen{+4.0}/\cgreen{+6.1}\\
\cmidrule(lr){2-9}
&ResNet101~\cite{he2016deep} & Convolution &  63.2 & 336.2 &40.0/36.1 &22.6/18.8 &44.0/39.7 &52.6/49.5 \\

\midrule
\multirow{5}{1cm}{HTC~\cite{chen2019hybrid}} 
&ResNet50~\cite{he2016deep} (\textit{baseline}) & Convolution & 80.0  &441.4&43.3/38.3  &24.3/19.9 &46.4/41.0 &57.7/53.0\\
&\textbf{DCDC-ResNet50} & \textbf{DCDC} &68.3 &368.7 &\textbf{46.1}/\textbf{40.6} &\textbf{28.9}/21.1 &49.2/43.2 &60.7/\textbf{58.8} \\
&\bm{$\Delta$} & &\cgreen{$\downarrow$ 15\%} & \cgreen{$\downarrow$ 16\%} &\cgreen{+2.8}/\cgreen{+2.3} &\cgreen{+4.6}/\cgreen{+1.2} &\cgreen{+2.8}/\cgreen{+2.2} &\cgreen{+3.0}/\cgreen{+5.8} \\
\cmidrule(lr){2-9}
&ResNet101~\cite{he2016deep}  & Convolution & 99.0  &517.5&44.8/39.6 & 25.7/21.3& 48.5/42.9& 60.2/55.0\\
&ResNeXt-101~\cite{xie2017aggregated} & Convolution &  98.7 &521.2& 46.1/40.5 & 27.1/\textbf{22.1} &\textbf{49.6}/\textbf{43.9} &\textbf{60.9}/55.5\\
\bottomrule
\end{tabular*}
\end{center}
\end{table*}

\begin{figure*}[t]
	\centering
	\includegraphics[width=18cm]{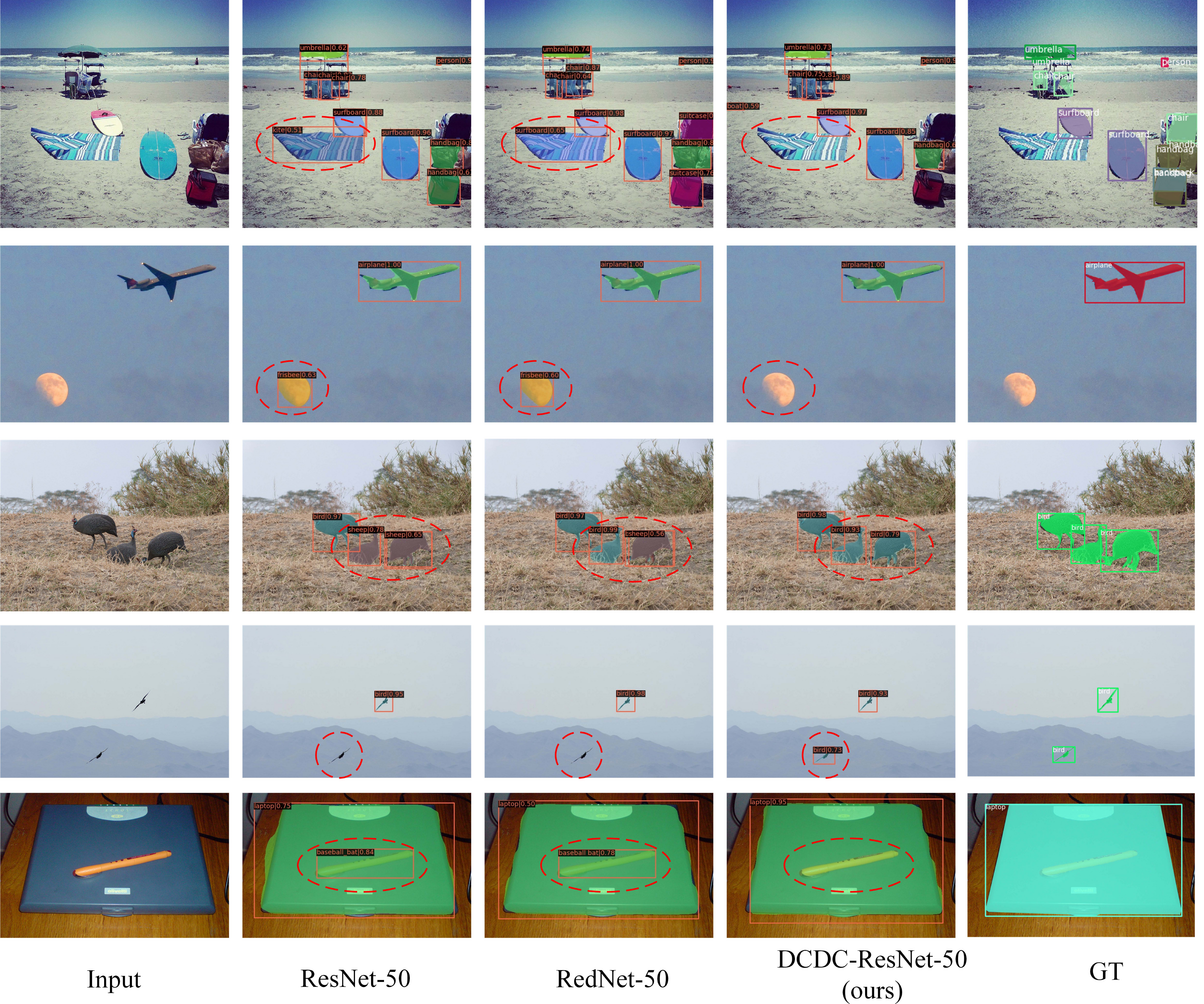}\\
	\caption{Some visualization examples of object detection and instance segmentation with Mask-RCNN benchmark~\cite{he2017mask} on COCO dataset. The comparative backbones include ResNet-50~\cite{he2016deep}, RedNet-50~\cite{li2021involution} and our proposed DCDC-ResNet-50. Here, GT means ground truth, and the results contained in the red dotted oval areas show the difference between our method and other competitors.}
	\label{detection_results}
\end{figure*}

\renewcommand{\arraystretch}{1.0}
\setlength\tabcolsep{1pt}
\begin{table*}\small
\caption{Performance comparisons of panoptic segmentation experiment on MS COCO dataset.}
\label{Seg_results}
\begin{center}
\begin{tabular*}{18 cm} {@{\extracolsep{\fill} }llccccccccccc}
\toprule
Segmentor & Backbone &Neck & $\#$Params/M& PQ & PQ$^\text{Th}$ & PQ$^\text{St}$ &SQ & SQ$^\text{Th}$ & SQ$^\text{St}$ &RQ & RQ$^\text{Th}$ & RQ$^\text{St}$\\
\midrule
 &ResNet50~\cite{he2016deep} (\textit{baseline})  & Convolution & 46 &40.3& 47.8& 29.0& 77.8& 80.9 & 73.2 & 49.4 & 57.6 & 37.0\\
 &RedNet50~\cite{li2021involution} & Convolution & 36 &42.4 & 49.3 & 32.0 & 78.4 & 81.0 & 74.5 & 51.8 & 59.3 & 40.5\\
 &\textbf{DCDC-ResNet50} & Convolution & 36.3 & 43.1 & 50.0 & 32.7 &78.6 & 81.1 & 74.9 & 52.6 & 60.1 & 41.2\\
 &\bm{$\Delta$}& & \cgreen{$\downarrow$ 21\%} &\cgreen{+2.8} & \cgreen{+2.2} & \cgreen{+3.7} & \cgreen{+0.8} & \cgreen{+0.2} & \cgreen{+1.7} &\cgreen{+3.2} &  \cgreen{+2.5} & \cgreen{+4.2}\\
{PFPN~\cite{kirillov2019panoptic_FPN}} &RedNet50~\cite{li2021involution} & Involution~\cite{li2021involution} & 34 & 42.8 & 49.8 & 32.1 & 78.4 & 80.8 & 74.8 & 52.3 & 60.2 & 40.4\\
 &\textbf{DCDC-ResNet50} & \textbf{DCDC} &  34.3 & \textbf{43.3} & \textbf{50.3}& \textbf{32.7} & \textbf{78.9} & 81.0 & \textbf{75.7} & \textbf{52.7} & \textbf{60.5} & \textbf{40.9}\\
  &\bm{$\Delta$}& &\cgreen{$\downarrow$ 25\%} &\cgreen{+3.0} & \cgreen{+2.5} & \cgreen{+3.7} & \cgreen{+1.1} & \cgreen{+0.1} & \cgreen{+2.5} &\cgreen{+3.3} &  \cgreen{+2.9} & \cgreen{+3.9}\\
\cmidrule(lr){2-13}
&ResNet101~\cite{he2016deep} &Convolution & 64.9 & 42.2 &50.1& 30.3& 78.3& \textbf{81.4}& 73.6 & 51.4 & 59.9 & 38.5\\
\bottomrule
\end{tabular*}
\end{center}
\end{table*}

\subsection{Object Detection and Instance Segmentation}
To verify the generalization of DCDC-ResNets on downstream compute vision tasks, we perform extensive object detection and instance segmentation experiments on MS COCO~\cite{lin2014microsoft}. We choose Faster RCNN~\cite{ren2015faster}, RetinaNet~\cite{lin2017focal}, and Mask RCNN~\cite{he2017mask} as the main benchmarks and conduct comparative experiments based on ResNet~\cite{he2016deep} backbones and the state-of-the-art dynamic convolutional network of RedNet~\cite{li2021involution}. In addition, we also perform experiments on superior benchmarks, such as HTC~\cite{chen2019hybrid}, to further verify the potential of DCDC-ResNets.

\subsubsection{Dataset and Evaluation Metrics}
The MS COCO dataset contains 118K training, 5K validation and 20K test images with 80 categories instances. For convenience, all experiments are evaluated on the validation set. We report COCO-style mean Average Precision of 
bounding boxes and masks  (i.e., AP$_\text{bbox}$ and AP$_\text{mask}$) with intersection over union (IoU) thresholds from 0.5 to 0.95 with an interval of 0.05. AP$_\text{S}$, AP$_\text{M}$, and AP$_\text{L}$ denote AP for small (pixel area $s \textless$32$^2$), medium (32$^2\leq s \leq$96$^2$), and large ($s \textgreater$96$^2$) objects, respectively.  

\subsubsection{Implementation Details}
Unless otherwise specified, we uniformly train all models for 12 epochs with stochastic gradient descent optimizer based on MMdetection library~\cite{2020mmclassification}. The initial learning rate, momentum and weight decay factors are set to 0.02 (0.01 for RetinaNet), 0.9, and 0.0001, respectively.
The learning rate is scaled by a factor of 0.1 at the 8-th and 11-th epochs. 
The 500 iterations linear warmup strategy of learning rate with a factor of 0.001 in first epoch is also adopted to guarantee smoothness of training processing.
The shorter and longer sides of input images are set to 800 and 1333. The mini-batchsize is set to 16 with eight GPUs. We carry out comparative experiments based on ResNet-50~\cite{he2016deep}, RedNet-50~\cite{li2021involution} and proposed DCDC-ResNet-50 backbones respectively. All backbones are pretrained on the ImageNet-1K~\cite{deng2009imagenet} dataset. For HTC, the total epoch is set to 20, and the learning rate is scaled by a factor of 0.1 at the 16-th and 19-th epochs.

\subsubsection{Main Results}
The performance comparisons are shown in Table~\ref{Det_results}. 
Overall, the proposed DCDC-ResNet enables all benchmarks to achieve significant performance improvements compared with the ResNet baseline and state-of-the-art dynamic convolution backbone of RedNet.
Specifically, compared with ResNet-50 based detectors, the DCDC-ResNet-50 based ones improve the performance by 2.7 AP and 3 AP, while reducing the $\#$Params and FLOPs by 23\%/14\% and 25\%/12\% on Faster RCNN~\cite{ren2015faster} and RetinaNet~\cite{lin2017focal} benchmarks, respectively. 
Furthermore, if we replace the convolutional layer in neck of detector with DCDC operator, we will get a more noticeable performance promotes of 3.2 AP, and the $\#$Params and FLOPs are further reduced by 28\% and 35\% on Faster RCNN~\cite{ren2015faster} benchmark. 
Compared with the RedNet-50~\cite{li2021involution}, our DCDC-ResNet-50 is still able to improve the performance by about 0.7$\sim$1.3 AP with almost no increase in $\#$Params and FLOPs on RetinaNet~\cite{lin2017focal} and Faster RCNN~\cite{ren2015faster} benchmarks.

For instance segmentation experiments with Mask RCNN~\cite{he2017mask},
compared with baseline ResNet-50, DCDC-ResNet-50 achieves 3.3 AP$_\text{bbox}$ and 2.0 AP$_\text{seg}$ improvements while reducing $\#$Params and FLOPs by 26\% and 28\%, respectively. 
DCDC-ResNet-50 outperforms dynamic convolutional network of RedNet-50 by 0.9 AP$_\text{bbox}$ and 0.7 AP$_\text{seg}$.
More gratifyingly, DCDC-ResNet-50 obtains 3.9 AP$_\text{seg}$ and 6.1 AP$_\text{seg}$ improvements for large scale object segmentation compared with RedNet-50 and ResNet-50, respectively.
This fully demonstrates the superiorities of the proposed designs of predicting LSA kernel via considering pixels in a wider range, and predicting GSI kernel via aggregating global features discussed in Sec.~\ref{SA_section} and Sec.~\ref{SI_section}, respectively.
Expermients based on HTC~\cite{chen2019hybrid} show that DCDC-ResNets are still able to bring the considerable performance gains on superior benchmarks.

\subsubsection{Visualization of Object Detection and Instance Segmentation}
In this subsection, we visualize some object detection and instance segmentation examples based on Mask-RCNN~\cite{he2017mask} benchmarks. As shown in Fig.~\ref{detection_results}, we randomly display some visualization example from evaluation split of the COCO dataset. It can be observed that the results based on proposed DCDC-ResNet-50 have lower false positive and higher recall rates compared with that based on ResNet-50 and RedNet-50, which demonstrates the effectiveness of the proposed methods.

\subsection{Panoptic Segmentation}
To further evaluate the adaptabilities of DCDC-ResNets, we extra conduct several experiments on the panoptic segmentation task. The more challenging panoptic segmentation task proposed in~\cite{kirillov2019panoptic} can be regarded as a combination of object instance and stuff semantic segmentation subtasks. 
We adopt panoptic feature pyramid networks (PFPN)~\cite{kirillov2019panoptic_FPN} as the benchmark.
Unless otherwise specified, the hyperparameters for all experiments are derived from MMdetection~\cite{2020mmclassification} and we train all models for 12 epochs on MS COCO panoptic segmentation dataset~\cite{lin2014microsoft}.

\subsubsection{Dataset and Evaluation Metrics}
In addition to the data of 80 categories things used in object detection and instance segmentation experiments, 53 categories labeled stuffs data are also included to train the models of panoptic segmentation task. All evaluations are performed on the validation set. We adopt standard evaluation metrics proposed in~\cite{kirillov2019panoptic} including PQ (panoptic quality), SQ (segmentation quality), and RQ (recognition quality), and report the performance on things and stuffs marked with \texttt{Th} and \texttt{St} superscripts, respectively.

\subsubsection{Main Results}
Table~\ref{Seg_results} compares the performances of the proposed DCDC-ResNet with those of ResNet~\cite{he2016deep} and RedNet~\cite{li2021involution}. It can be observed that the DCDC-ResNet-50 can remarkably achieve 2.8 PQ, 0.8 SQ and 3.2 RQ improvements respectively compared with ResNet-50 baseline, while costing 21\% fewer $\#$Params. 
If we replace all convolution layers in the neck of framework with DCDC operator, 3.0 PQ, 1.1 SQ and 3.3 RQ boosts can be further achieved. 
In addition, compared with the state-of-the-art dynamic convolutional backbone of RedNet-50, our methods still obtain  0.7 PQ, 0.2 SQ and 0.8 RQ improvements with almost no increase of parameters. 
Note that the experimental results with DCDC-ResNet-50 exceed that with ResNet-101 (growths of 1.1 PQ, 0.6 SQ and 1.3 RQ), while the $\#$Params is only about half of it (34.3 M \textit{vs.} 64.9 M).

\section{Conclusion}
In this paper, we novelly model the features as a combination of local spatial-adaptive and global shift-invariant parts, and then propose a two-branch dual complementary dynamic convolution operator to properly deal with these two types of features, which significantly enhance the representation capacity. The DCDC-ResNets built based on the proposed operator achieves obviously better performance than ResNet baseline and most of the state-of-the-art dynamic convolutional networks, while with fewer number of parameters and FLOPs. 
We also perform transfer experiments on downstream vision tasks including object detection, instance and panoptic segmentations to evaluate the generalization ability of model, the experimental results have shown remarkable performance improvements.

\bibliographystyle{IEEEtran}
\bibliography{egbib}

\begin{thebibliography}{10}
\providecommand{\url}[1]{#1}
\csname url@samestyle\endcsname
\providecommand{\newblock}{\relax}
\providecommand{\bibinfo}[2]{#2}
\providecommand{\BIBentrySTDinterwordspacing}{\spaceskip=0pt\relax}
\providecommand{\BIBentryALTinterwordstretchfactor}{4}
\providecommand{\BIBentryALTinterwordspacing}{\spaceskip=\fontdimen2\font plus
\BIBentryALTinterwordstretchfactor\fontdimen3\font minus
  \fontdimen4\font\relax}
\providecommand{\BIBforeignlanguage}[2]{{%
\expandafter\ifx\csname l@#1\endcsname\relax
\typeout{** WARNING: IEEEtran.bst: No hyphenation pattern has been}%
\typeout{** loaded for the language `#1'. Using the pattern for}%
\typeout{** the default language instead.}%
\else
\language=\csname l@#1\endcsname
\fi
#2}}
\providecommand{\BIBdecl}{\relax}
\BIBdecl

\bibitem{lecun1998gradient}
Y.~LeCun, L.~Bottou, Y.~Bengio, and P.~Haffner, ``Gradient-based learning
  applied to document recognition,'' \emph{Proceedings of the IEEE}, vol.~86,
  no.~11, pp. 2278--2324, 1998.

\bibitem{krizhevsky2012imagenet}
A.~Krizhevsky, I.~Sutskever, and G.~E. Hinton, ``Imagenet classification with
  deep convolutional neural networks,'' \emph{Adv. Neural Inform. Process.
  Syst.}, vol.~25, pp. 1097--1105, 2012.

\bibitem{he2016deep}
K.~He, X.~Zhang, S.~Ren, and J.~Sun, ``Deep residual learning for image
  recognition,'' in \emph{IEEE Conf. Comput. Vis. Pattern Recognit.}, 2016, pp.
  770--778.

\bibitem{ren2015faster}
S.~Ren, K.~He, R.~Girshick, and J.~Sun, ``Faster r-cnn: Towards real-time
  object detection with region proposal networks,'' \emph{Adv. Neural Inform.
  Process. Syst.}, vol.~28, pp. 91--99, 2015.

\bibitem{chen2020high}
X.~Chen, H.~Li, Q.~Wu, K.~N. Ngan, and L.~Xu, ``High-quality r-cnn object
  detection using multi-path detection calibration network,'' \emph{IEEE
  Transactions on Circuits and Systems for Video Technology}, vol.~31, no.~2,
  pp. 715--727, 2020.

\bibitem{zheng2021multi}
Y.~Zheng, X.~Liu, B.~Xiao, X.~Cheng, Y.~Wu, and S.~Chen, ``Multi-task
  convolution operators with object detection for visual tracking,'' \emph{IEEE
  Trans. Circuit Syst. Video Technol.}, 2021.

\bibitem{he2017mask}
K.~He, G.~Gkioxari, P.~Doll{\'a}r, and R.~Girshick, ``Mask r-cnn,'' in
  \emph{IEEE Int. Conf. Comput. Vis.}, 2017, pp. 2961--2969.

\bibitem{sun2021gaussian}
X.~Sun, C.~Chen, X.~Wang, J.~Dong, H.~Zhou, and S.~Chen, ``Gaussian dynamic
  convolution for efficient single-image segmentation,'' \emph{IEEE Trans.
  Circuit Syst. Video Technol.}, vol.~32, no.~5, pp. 2937--2948, 2021.

\bibitem{ding2021dynamic}
Y.~Ding, Z.~Han, Y.~Zhou, Y.~Zhu, J.~Chen, Q.~Ye, and J.~Jiao, ``Dynamic
  perception framework for fine-grained recognition,'' \emph{IEEE Trans.
  Circuit Syst. Video Technol.}, vol.~32, no.~3, pp. 1353--1365, 2021.

\bibitem{yang2019condconv}
B.~Yang, G.~Bender, Q.~V. Le, and J.~Ngiam, ``Condconv: conditionally
  parameterized convolutions for efficient inference,'' in \emph{Adv. Neural
  Inform. Process. Syst.}, 2019, pp. 1307--1318.

\bibitem{chen2020dynamic}
Y.~Chen, X.~Dai, M.~Liu, D.~Chen, L.~Yuan, and Z.~Liu, ``Dynamic convolution:
  Attention over convolution kernels,'' in \emph{IEEE Conf. Comput. Vis.
  Pattern Recognit.}, 2020, pp. 11\,030--11\,039.

\bibitem{zhang2020dynet}
Y.~Zhang, J.~Zhang, Q.~Wang, and Z.~Zhong, ``Dynet: Dynamic convolution for
  accelerating convolutional neural networks,'' \emph{arXiv preprint
  arXiv:2004.10694}, 2020.

\bibitem{li2021involution}
D.~Li, J.~Hu, C.~Wang, X.~Li, Q.~She, L.~Zhu, T.~Zhang, and Q.~Chen,
  ``Involution: Inverting the inherence of convolution for visual
  recognition,'' in \emph{IEEE Conf. Comput. Vis. Pattern Recognit.}, 2021, pp.
  12\,321--12\,330.

\bibitem{zhou2021decoupled}
J.~Zhou, V.~Jampani, Z.~Pi, Q.~Liu, and M.-H. Yang, ``Decoupled dynamic filter
  networks,'' in \emph{IEEE Conf. Comput. Vis. Pattern Recognit.}, 2021, pp.
  6647--6656.

\bibitem{xie2017aggregated}
S.~Xie, R.~Girshick, P.~Doll{\'a}r, Z.~Tu, and K.~He, ``Aggregated residual
  transformations for deep neural networks,'' in \emph{IEEE Conf. Comput. Vis.
  Pattern Recognit.}, 2017, pp. 1492--1500.

\bibitem{howard2017mobilenets}
A.~G. Howard, M.~Zhu, B.~Chen, D.~Kalenichenko, W.~Wang, T.~Weyand,
  M.~Andreetto, and H.~Adam, ``Mobilenets: Efficient convolutional neural
  networks for mobile vision applications,'' in \emph{IEEE Conf. Comput. Vis.
  Pattern Recognit.}, 2017.

\bibitem{sandler2018mobilenetv2}
M.~Sandler, A.~Howard, M.~Zhu, A.~Zhmoginov, and L.-C. Chen, ``Mobilenetv2:
  Inverted residuals and linear bottlenecks,'' in \emph{IEEE Conf. Comput. Vis.
  Pattern Recognit.}, 2018, pp. 4510--4520.

\bibitem{lin2013network}
M.~Lin, Q.~Chen, and S.~Yan, ``Network in network,'' \emph{arXiv preprint
  arXiv:1312.4400}, 2013.

\bibitem{yu2015multi}
F.~Yu and V.~Koltun, ``Multi-scale context aggregation by dilated
  convolutions,'' in \emph{Int. Conf. Learn. Represent.}, 2016.

\bibitem{liang2015semantic}
C.~Liang-Chieh, G.~Papandreou, I.~Kokkinos, K.~Murphy, and A.~Yuille,
  ``Semantic image segmentation with deep convolutional nets and fully
  connected crfs,'' in \emph{Int. Conf. Learn. Represent.}, 2015.

\bibitem{chen2017deeplab}
L.-C. Chen, G.~Papandreou, I.~Kokkinos, K.~Murphy, and A.~L. Yuille, ``Deeplab:
  Semantic image segmentation with deep convolutional nets, atrous convolution,
  and fully connected crfs,'' \emph{IEEE Trans. Pattern Anal. Mach. Intell.},
  vol.~40, no.~4, pp. 834--848, 2017.

\bibitem{han2021dynamic}
Y.~Han, G.~Huang, S.~Song, L.~Yang, H.~Wang, and Y.~Wang, ``Dynamic neural
  networks: A survey,'' \emph{arXiv preprint arXiv:2102.04906}, 2021.

\bibitem{huang2018multi}
G.~Huang, D.~Chen, T.~Li, F.~Wu, L.~van~der Maaten, and K.~Weinberger,
  ``Multi-scale dense networks for resource efficient image classification,''
  in \emph{Int. Conf. Learn. Represent.}, 2018.

\bibitem{bolukbasi2017adaptive}
T.~Bolukbasi, J.~Wang, O.~Dekel, and V.~Saligrama, ``Adaptive neural networks
  for efficient inference,'' in \emph{Int. Conf. Machine Learning}.\hskip 1em
  plus 0.5em minus 0.4em\relax PMLR, 2017, pp. 527--536.

\bibitem{eigen2013learning}
D.~Eigen, M.~Ranzato, and I.~Sutskever, ``Learning factored representations in
  a deep mixture of experts,'' \emph{arXiv preprint arXiv:1312.4314}, 2013.

\bibitem{bengio2015conditional}
E.~Bengio, P.-L. Bacon, J.~Pineau, and D.~Precup, ``Conditional computation in
  neural networks for faster models,'' \emph{arXiv preprint arXiv:1511.06297},
  2015.

\bibitem{wang2018skipnet}
X.~Wang, F.~Yu, Z.-Y. Dou, T.~Darrell, and J.~E. Gonzalez, ``Skipnet: Learning
  dynamic routing in convolutional networks,'' in \emph{Eur. Conf. Comput.
  Vis.}, 2018, pp. 409--424.

\bibitem{veit2018convolutional}
A.~Veit and S.~Belongie, ``Convolutional networks with adaptive inference
  graphs,'' in \emph{Eur. Conf. Comput. Vis.}, 2018, pp. 3--18.

\bibitem{ma2020weightnet}
N.~Ma, X.~Zhang, J.~Huang, and J.~Sun, ``Weightnet: Revisiting the design space
  of weight networks,'' in \emph{Eur. Conf. Comput. Vis.}\hskip 1em plus 0.5em
  minus 0.4em\relax Springer, 2020, pp. 776--792.

\bibitem{li2020revisiting}
Y.~Li, Y.~Chen, X.~Dai, D.~Chen, Y.~Yu, L.~Yuan, Z.~Liu, M.~Chen,
  N.~Vasconcelos \emph{et~al.}, ``Revisiting dynamic convolution via matrix
  decomposition,'' in \emph{Int. Conf. Learn. Represent.}, 2020.

\bibitem{jia2016dynamic}
X.~Jia, B.~De~Brabandere, T.~Tuytelaars, and L.~V. Gool, ``Dynamic filter
  networks,'' \emph{Adv. Neural Inform. Process. Syst.}, vol.~29, pp. 667--675,
  2016.

\bibitem{harley2017segmentation}
A.~W. Harley, K.~G. Derpanis, and I.~Kokkinos, ``Segmentation-aware
  convolutional networks using local attention masks,'' in \emph{IEEE Int.
  Conf. Comput. Vis.}, 2017, pp. 5038--5047.

\bibitem{su2019pixel}
H.~Su, V.~Jampani, D.~Sun, O.~Gallo, E.~Learned-Miller, and J.~Kautz,
  ``Pixel-adaptive convolutional neural networks,'' in \emph{IEEE Conf. Comput.
  Vis. Pattern Recognit.}, 2019, pp. 11\,166--11\,175.

\bibitem{denil2013predicting}
M.~Denil, B.~Shakibi, L.~Dinh, M.~Ranzato, and N.~De~Freitas, ``Predicting
  parameters in deep learning,'' in \emph{Adv. Neural Inform. Process. Syst.},
  2013.

\bibitem{ha2016hypernetworks}
D.~Ha, A.~Dai, and Q.~V. Le, ``Hypernetworks,'' in \emph{Int. Conf. Learn.
  Represent.}, 2016.

\bibitem{jaderberg2015spatial}
M.~Jaderberg, K.~Simonyan, A.~Zisserman \emph{et~al.}, ``Spatial transformer
  networks,'' \emph{Adv. Neural Inform. Process. Syst.}, vol.~28, pp.
  2017--2025, 2015.

\bibitem{dai2017deformable}
J.~Dai, H.~Qi, Y.~Xiong, Y.~Li, G.~Zhang, H.~Hu, and Y.~Wei, ``Deformable
  convolutional networks,'' in \emph{IEEE Int. Conf. Comput. Vis.}, 2017, pp.
  764--773.

\bibitem{zhu2019deformable}
X.~Zhu, H.~Hu, S.~Lin, and J.~Dai, ``Deformable convnets v2: More deformable,
  better results,'' in \emph{IEEE Conf. Comput. Vis. Pattern Recognit.}, 2019,
  pp. 9308--9316.

\bibitem{zamora2019adaptive}
J.~Zamora~Esquivel, A.~Cruz~Vargas, P.~Lopez~Meyer, and O.~Tickoo, ``Adaptive
  convolutional kernels,'' in \emph{IEEE Int. Conf. Comput. Vis. Worksh.},
  2019, pp. 0--0.

\bibitem{wang2019carafe}
J.~Wang, K.~Chen, R.~Xu, Z.~Liu, C.~C. Loy, and D.~Lin, ``{CARAFE}:
  Content-aware reassembly of features,'' in \emph{IEEE Int. Conf. Comput.
  Vis.}, 2019, pp. 3007--3016.

\bibitem{gao2019lip}
Z.~Gao, L.~Wang, and G.~Wu, ``Lip: Local importance-based pooling,'' in
  \emph{IEEE Int. Conf. Comput. Vis.}, 2019, pp. 3355--3364.

\bibitem{wang2021carafe++}
J.~Wang, K.~Chen, R.~Xu, Z.~Liu, C.~C. Loy, and D.~Lin, ``{CARAFE++}: Unified
  content-aware reassembly of features,'' \emph{IEEE Trans. Pattern Anal. Mach.
  Intell.}, 2021.

\bibitem{hu2018squeeze}
J.~Hu, L.~Shen, and G.~Sun, ``Squeeze-and-excitation networks,'' in \emph{IEEE
  Conf. Comput. Vis. Pattern Recognit.}, 2018, pp. 7132--7141.

\bibitem{deng2009imagenet}
J.~Deng, W.~Dong, R.~Socher, L.-J. Li, K.~Li, and L.~Fei-Fei, ``Imagenet: A
  large-scale hierarchical image database,'' in \emph{IEEE Conf. Comput. Vis.
  Pattern Recognit.}\hskip 1em plus 0.5em minus 0.4em\relax IEEE, 2009, pp.
  248--255.

\bibitem{lin2014microsoft}
T.-Y. Lin, M.~Maire, S.~Belongie, J.~Hays, P.~Perona, D.~Ramanan,
  P.~Doll{\'a}r, and C.~L. Zitnick, ``Microsoft coco: Common objects in
  context,'' in \emph{Eur. Conf. Comput. Vis.}\hskip 1em plus 0.5em minus
  0.4em\relax Springer, 2014, pp. 740--755.

\bibitem{mmdetection}
K.~Chen, J.~Wang, J.~Pang, Y.~Cao, Y.~Xiong, X.~Li, S.~Sun, W.~Feng, Z.~Liu,
  J.~Xu, Z.~Zhang, D.~Cheng, C.~Zhu, T.~Cheng, Q.~Zhao, B.~Li, X.~Lu, R.~Zhu,
  Y.~Wu, J.~Dai, J.~Wang, J.~Shi, W.~Ouyang, C.~C. Loy, and D.~Lin,
  ``{MMDetection}: Open mmlab detection toolbox and benchmark,'' \emph{arXiv
  preprint arXiv:1906.07155}, 2019.

\bibitem{rw2019timm}
R.~Wightman, ``Pytorch image models,''
  \url{https://github.com/rwightman/pytorch-image-models}, 2019.

\bibitem{hu2019local}
H.~Hu, Z.~Zhang, Z.~Xie, and S.~Lin, ``Local relation networks for image
  recognition,'' in \emph{IEEE Int. Conf. Comput. Vis.}, 2019, pp. 3464--3473.

\bibitem{ramachandran2019stand}
P.~Ramachandran, N.~Parmar, A.~Vaswani, I.~Bello, A.~Levskaya, and J.~Shlens,
  ``Stand-alone self-attention in vision models,'' \emph{Adv. Neural Inform.
  Process. Syst.}, vol.~32, 2019.

\bibitem{zhao2020exploring}
H.~Zhao, J.~Jia, and V.~Koltun, ``Exploring self-attention for image
  recognition,'' in \emph{IEEE Conf. Comput. Vis. Pattern Recognit.}, 2020, pp.
  10\,076--10\,085.

\bibitem{woo2018cbam}
S.~Woo, J.~Park, J.-Y. Lee, and I.~S. Kweon, ``Cbam: Convolutional block
  attention module,'' in \emph{Eur. Conf. Comput. Vis.}, 2018, pp. 3--19.

\bibitem{bello2019attention}
I.~Bello, B.~Zoph, A.~Vaswani, J.~Shlens, and Q.~V. Le, ``Attention augmented
  convolutional networks,'' in \emph{IEEE Int. Conf. Comput. Vis.}, 2019, pp.
  3286--3295.

\bibitem{wang2020axial}
H.~Wang, Y.~Zhu, B.~Green, H.~Adam, A.~Yuille, and L.-C. Chen, ``Axial-deeplab:
  Stand-alone axial-attention for panoptic segmentation,'' in \emph{Eur. Conf.
  Comput. Vis.}\hskip 1em plus 0.5em minus 0.4em\relax Springer, 2020, pp.
  108--126.

\bibitem{gao2019res2net}
S.~Gao, M.-M. Cheng, K.~Zhao, X.-Y. Zhang, M.-H. Yang, and P.~H. Torr,
  ``Res2net: A new multi-scale backbone architecture,'' \emph{IEEE Trans.
  Pattern Anal. Mach. Intell.}, 2019.

\bibitem{li2021omni}
C.~Li, A.~Zhou, and A.~Yao, ``Omni-dimensional dynamic convolution,'' in
  \emph{Int. Conf. Learn. Represent.}, 2021.

\bibitem{selvaraju2017grad}
R.~R. Selvaraju, M.~Cogswell, A.~Das, R.~Vedantam, D.~Parikh, and D.~Batra,
  ``Grad-cam: Visual explanations from deep networks via gradient-based
  localization,'' in \emph{IEEE Int. Conf. Comput. Vis.}, 2017, pp. 618--626.

\bibitem{lin2017focal}
T.-Y. Lin, P.~Goyal, R.~Girshick, K.~He, and P.~Doll{\'a}r, ``Focal loss for
  dense object detection,'' in \emph{IEEE Int. Conf. Comput. Vis.}, 2017, pp.
  2980--2988.

\bibitem{chen2019hybrid}
K.~Chen, J.~Pang, J.~Wang, Y.~Xiong, X.~Li, S.~Sun, W.~Feng, Z.~Liu, J.~Shi,
  W.~Ouyang \emph{et~al.}, ``Hybrid task cascade for instance segmentation,''
  in \emph{IEEE Conf. Comput. Vis. Pattern Recognit.}, 2019, pp. 4974--4983.

\bibitem{kirillov2019panoptic_FPN}
A.~Kirillov, R.~Girshick, K.~He, and P.~Doll{\'a}r, ``Panoptic feature pyramid
  networks,'' in \emph{IEEE Conf. Comput. Vis. Pattern Recognit.}, 2019, pp.
  6399--6408.

\bibitem{2020mmclassification}
M.~Contributors, ``Openmmlab's image classification toolbox and benchmark,''
  \url{https://github.com/open-mmlab/mmclassification}, 2020.

\bibitem{kirillov2019panoptic}
A.~Kirillov, K.~He, R.~Girshick, C.~Rother, and P.~Doll{\'a}r, ``Panoptic
  segmentation,'' in \emph{IEEE Conf. Comput. Vis. Pattern Recognit.}, 2019,
  pp. 9404--9413.

\end{thebibliography}

\end{document}